\documentclass[journal]{IEEEtran} 

\usepackage{dblfloatfix}
\usepackage[numbers, sectionbib, sort]{natbib}
\usepackage{bm}
\usepackage{gensymb}
\usepackage{xcolor}
\usepackage{graphicx}
\usepackage{amsmath}
\usepackage{amssymb}
\usepackage{subcaption}
\usepackage{amsfonts}
\usepackage{siunitx}
\usepackage{booktabs}
\usepackage{makecell}
\usepackage{multirow}
\usepackage{upgreek}
\usepackage[font=small]{caption}
\usepackage[export]{adjustbox}
\usepackage{tikz}
\usepackage{tabularx}
\usepackage{sidecap} \sidecaptionvpos{figure}{c}
\usepackage[hidelinks]{hyperref}
\usepackage[nameinlink, capitalize]{cleveref}
\usepackage[printonlyused,withpage,nolist,nohyperlinks]{acronym}

\Crefname{section}{Sec.}{Sec.}
\Crefname{equation}{Eq.}{Eq.}

\usepackage{soul}

\newcommand{\etal}{\textit{et al}.}

\def\etalcite#1{\etal~\cite{#1}}

\IEEEoverridecommandlockouts

\title{Efficient View Planning Guided by Previous-Session Reconstruction for Repeated Plant Monitoring}

\author{Sicong Pan, Luca Lobefaro, Moein Taherkhani, Xuying Huang, Rohit Menon, Cyrill Stachniss, Maren Bennewitz %
\thanks{S. Pan, M. Taherkhani, X. Huang, R. Menon, and M. Bennewitz are with the Humanoid Robots Lab, University of Bonn. L. Lobefaro and C. Stachniss are with the Lab for Photogrammetry and Robotics, University of Bonn. M. Bennewitz and C. Stachniss are additionally with the Lamarr Institute for Machine Learning and Artificial Intelligence, and the Center for Robotics, University of Bonn, Germany.
This work has partially been funded by the Deutsche Forschungsgemeinschaft (DFG, German Research Foundation) under grant 459376902 – AID4Crops, under Germany’s Excellence Strategy, EXC-2070 – 390732324 – PhenoRob, and by the German Federal Ministry of Research, Technology and Space (BMFTR) under the Robotics Institute Germany (RIG), grant No. 16ME0999.
}
\vspace{-0.5cm}
}

\begin{document}

\maketitle
\thispagestyle{empty}
\pagestyle{empty}
\begin{abstract}
Repeated plant monitoring is essential for tracking crop growth, and 3D reconstruction enables consistent comparison across monitoring sessions.
However, rebuilding a 3D model from scratch in every session is costly and ignores information observed previously.
We propose efficient view planning guided by a previous-session reconstruction, which reuses a 3D model from the previous session to improve active perception in the current session.
Based on this previous-session reconstruction, our method replaces iterative next-best-view planning with one-shot view planning that selects an informative set of views and computes the shortest execution path connecting them under a defined motion-cost model.
Experiments on real multi-session datasets, including public single-plant scans and a newly collected greenhouse crop-row dataset, show that our method achieves comparable or higher surface coverage with fewer executed views and shorter robot paths than iterative and one-shot baselines.
Real-world greenhouse experiments further demonstrate the practical deployability and effectiveness of our method for repeated plant monitoring.
\end{abstract}

\section{Introduction}

Monitoring crop growth is essential for understanding plant development and supporting timely agricultural decision-making~\cite{yang2020crop}, yet it remains labor-intensive.
To enable consistent comparisons across sessions, monitoring systems increasingly rely on 3D reconstruction of plant geometry.
Autonomous monitoring with robots can reduce manual effort, and recent work has explored active perception for 3D plant reconstruction by solving next-best-view (NBV) planning problems~\citep{zaenker2021viewpoint,menon2023iros,burusa2024attention,ci2025ssl} to reduce the impact of occlusions caused by complex plant structures.
In production environments such as greenhouses, 3D reconstruction is carried out repeatedly across monitoring sessions (e.g., multiple times per week) to track continuous growth.
However, many existing systems treat each session independently and restart the reconstruction pipeline, leading to redundant data acquisition and discarding informative geometry already observed in previous sessions.

\begin{figure}[!t]
\centering
\includegraphics[width=1.0\columnwidth]{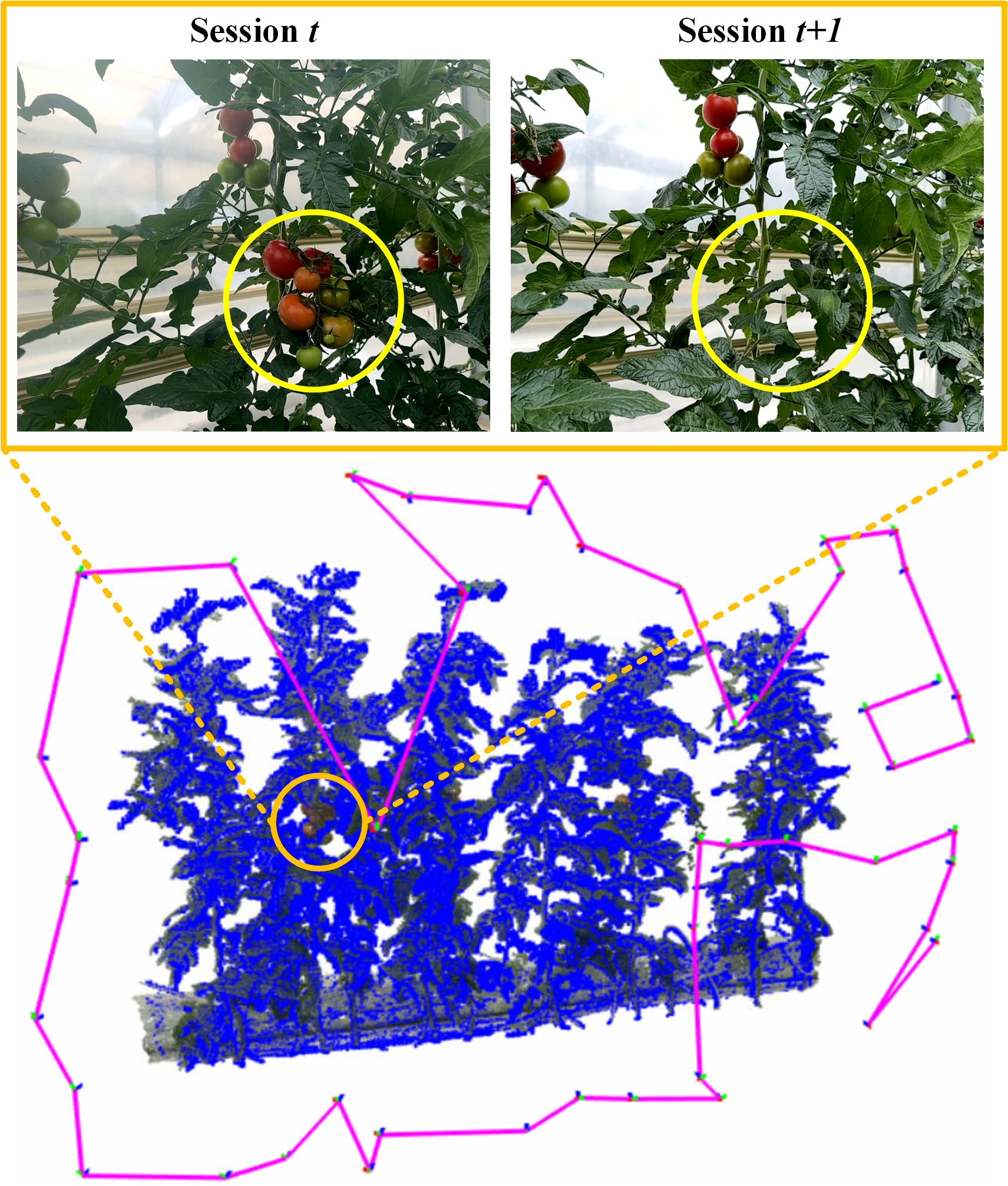}
\caption{
Example of view planning guided by a previous-session reconstruction in the greenhouse-row setting.
Top: Representative RGB views from two monitoring sessions ($t$ and $t{+}1$) illustrate session-to-session changes in foliage and fruit configuration (yellow circles), e.g., due to plant growth and routine harvesting.
Bottom: We first build a coarse reconstruction (blue) for the current session and then align the previous-session reconstruction to obtain an actionable approximation (colored) that highlights under-observed structures and guides planning toward poorly covered regions.
Based on this approximation, the proposed method selects informative views and computes the shortest execution path connecting them (purple) for efficient data acquisition with an eye-in-hand camera on a robotic arm using a greenhouse mobile robot platform~\cite{smitt2021pathobot}.
}
\label{fig_cover}
\end{figure}

In this paper, we ask whether a 3D reconstruction from a previous monitoring session can be exploited to make view planning for the current session more efficient.
We propose an efficient view-planning approach guided by the previous-session reconstruction, with the goal of reducing acquisition cost (executed views and robot path lengths) while maintaining high reconstruction quality.
Our motivation is that many iterative next-best-view (NBV) methods select views greedily and optimize them locally, which can lead to unnecessary view executions when a global execution plan is required.
In contrast, one-shot view planning~\cite{pan2022ral1,pan2024tro} selects an informative set of views at once by solving a set covering optimization problem and then optimizes their visitation order through a traveling salesman problem (TSP), offering a more acquisition-efficient alternative.
However, solving the set covering optimization problem relies on an approximate geometric representation of the current scene.
To obtain such a representation in repeated monitoring, we transfer geometry from the previous session: we align the previous-session reconstruction to a coarse current-session reconstruction using non-rigid registration~\cite{sumner2007embedded,lobefaro2024spatio}, yielding a coarse yet actionable approximation of the current plant geometry.
Based on this approximation, our method reduces redundant observations and shortens planned execution paths compared with iterative NBV baselines, while preserving reconstruction quality.

In particular, this non-rigid registration-based transfer step requires a sufficiently accurate coarse reconstruction of the current session.
To meet this warm-start requirement, we adopt different strategies at two reconstruction scales: individual plants and greenhouse crop rows.
For individual plants, we first execute a short NBV stage after initial observation to obtain a coarse reconstruction.
Adding one NBV can substantially increase surface coverage (as observed in~\citep{pan2024tro}), providing a reliable basis for registration.
For greenhouse rows, we use a passive mapping procedure~\citep{smitt2021pathobot} to build the coarse reconstruction: the mobile robot platform performs a single sweep on the rail with three fixed lateral cameras, providing a coarse reconstruction without active perception.

Over several days between monitoring sessions, new plant structures may emerge beyond the previously reconstructed geometry.
These warm-starts improve non-rigid alignment robustness, but they may still underestimate the true surface due to inter-session growth and residual registration errors.
Since our one-shot planner performs set cover on the candidate unobserved surface, such underestimation can exclude regions that have newly appeared in the current session, resulting in insufficient viewpoint coverage.
We therefore augment the transferred planning geometry with a conservative neighborhood around current observations in regions insufficiently represented by the aligned previous-session reconstruction.
This adds a spatial tolerance around newly observed or imperfectly aligned structures, producing conservative planning geometry for subsequent view selection.
Fig.~\ref{fig_cover} illustrates our approach in the greenhouse crop row setting.

We further adapt the view parameterization to the geometric characteristics of the two reconstruction settings.
For individual plants, we reduce 6D view planning to selecting 3D camera positions while fixing the look-at direction toward the plant centroid, as commonly done in object-centric view planning~\cite{pan2024tro}.
This simplification retains reconstruction coverage while reducing the search space.
In contrast, greenhouse crop-row monitoring operates at the meter scale with restricted fields of view. 
As a result, effective coverage requires jointly planning both 3D camera positions and viewing directions (i.e., 3D look-at targets).

To the best of our knowledge, this is the first work to address the task of repeated plant monitoring with view planning guided by a previous-session reconstruction, across both individual plants and greenhouse crop rows under severe self-occlusion.
Our contributions are:
\begin{itemize}
    \item \textbf{Efficient view planning for repeated plant monitoring.} We propose a one-shot view-planning approach that selects an informative \emph{set} of views while explicitly accounting for execution effort (e.g., path length), enabled by guidance from a previous-session reconstruction.
    \item \textbf{Robust previous-session reconstruction transfer.} We build an actionable planning geometry by (i)~obtaining a coarse current-session reconstruction for warm-start, (ii)~aligning the previous-session reconstruction via non-rigid registration, and (iii)~applying a conservative inflation to improve robustness to inter-session growth and residual registration errors.
    \item \textbf{Setting-dependent view parameterization and use of transferred geometry.} We tailor the planning variables to the monitoring setting: for individual plants, we reduce 6D planning to 3D position selection by fixing look-at toward the plant centroid; for greenhouse crop rows, we jointly plan 3D positions and look-at directions, where the transferred geometry primarily guides look-at selection toward previously under-observed regions.
\end{itemize}

We conduct experiments on real plants captured over repeated monitoring sessions, including public single-plant scans~\cite{chebrolu2021registration} and a newly collected greenhouse crop-row dataset\footnote{\url{www.kaggle.com/datasets/sicongpan/greenhouse-multi-session-row-dataset}}.
We further validate the method on a greenhouse robotic platform, incorporating robot-specific execution constraints, joint-space route ordering, and CUDA-accelerated visibility evaluation for practical deployment.
To facilitate reproducibility, we release our implementation on Github\footnote{\url{www.github.com/HumanoidsBonn/TPVP}}.

\section{Related Work}

\subsection{View Planning in 3D Reconstruction}
3D reconstruction is central to many robotic applications, including inspection, mapping, and manipulation.
Without a reliable geometric prior, a common paradigm is NBV planning: the robot iteratively selects the next camera pose based on the current partial reconstruction to greedily maximize an information-gain objective, using either search-based~\cite{zaenker2021viewpoint,menon2023iros,burusa2024attention,pan2022ral2,burusa2024gradient} or learning-based~\cite{ci2025ssl,zeng2020iros,mendoza2020prl,la2025enhancing,yi2025view} techniques.
This paradigm has also been extended to agricultural environments, where dense foliage and severe self-occlusion make perception particularly challenging. 
As an example, Zaenker~\etalcite{zaenker2021viewpoint} applied NBV planning to fruit size and position estimation in greenhouse plants.
Their method builds an octree representation with labeled fruit regions of interest (ROIs) and uses it to alternate between ROI-targeted viewpoint sampling and more general exploration, thereby improving observation coverage under occlusion.
Their NBV approach requires updating the map, re-evaluating candidate views, and recomputing visibility or uncertainty estimates, which tightly couple planning with online perception and incur substantial computational cost. 
Moreover, greedy per-step decisions are prone to local optima: they may continue refining already well-observed regions while postponing harder-to-observe structures, leading to redundant back-and-forth motions and inefficient view sequences.

\begin{figure*}[!t]
\centering
\includegraphics[width=1.0\textwidth]{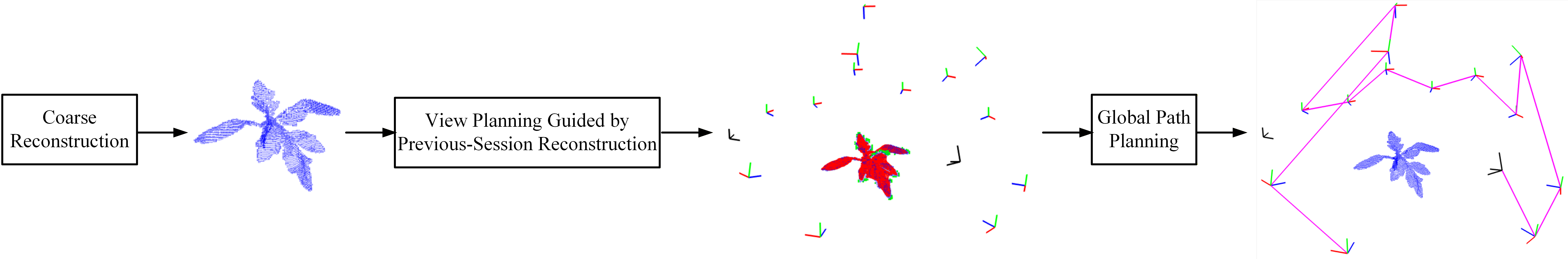}
\caption{
Overview of the proposed one-shot view planning pipeline.
Shown on an individual-plant example, the approach starts from a coarse reconstruction (blue); in this example, it is obtained from an initial view fused with one warm-up NBV observation (black).
We then transfer the previous-session reconstruction (red) to obtain actionable geometric guidance and select a minimal view set (colored axes) that covers the plant.
Finally, the selected views are ordered by global path planning, yielding a global execution path (purple) that reduces travel during data collection.
}
\label{fig_pipeline}
\vspace{-0.2cm}
\end{figure*}

To mitigate these limitations, one-shot view planning~\cite{pan2022ral1,hu2024icra,pan2024icra,pan2024iros,pan2025dm,patino2026one} selects a set of informative views from an initial measurement and then computes a globally short path through them, reducing online replanning overhead and improving travel efficiency.
Recent systems further combine NBV with one-shot planning~\cite{pan2024tro}: a short NBV warm-up increases initial observability and then triggers set-cover-style view selection to preserve reconstruction quality while keeping trajectories efficient.
However, existing pipelines often rely on learned shape priors trained on generic object datasets, which are not tailored to agricultural scenes and do not capture plant-specific, time-varying geometry.
Our approach replaces generic learned shape priors with plant-specific guidance from a previous-session reconstruction: we reuse the reconstruction of the same plant from the previous monitoring session to guide view and look-at selection within a one-shot planning pipeline, preserving global path efficiency while accounting for session-to-session changes.

Relatedly, global view-motion planning has been studied for fruit mapping~\cite{zaenker2023graph,jose2025go}, where view selection and robot motion are jointly optimized to cover task-relevant fruit regions based on detections.
In contrast, repeated plant monitoring requires structure-level reconstruction beyond fruit regions (e.g., canopy geometry, occluded leaves, and growth-induced deformations), since many monitoring signals are encoded in the overall plant shape rather than sparse targets.
Accordingly, planning for surface completeness benefits from a geometric representation that supports coverage evaluation and occlusion reasoning over the full plant.
This is the setting we address: view planning for repeated plant monitoring guided by a previous-session reconstruction, which provides plant-specific geometric guidance for efficient data acquisition in the current session.

\subsection{Non-Rigid Registration}
3D plant reconstruction poses additional challenges due to plant growth and motion, especially when integrating recordings from different sessions.
Non-rigid registration solves this by deforming a source point cloud or mesh to align with a target shape.

Non-rigid registration is a well-studied problem in computer graphics~\cite{deng2022survey}.
Most methods extend the rigid iterative closest point (ICP)~\cite{besl1992pami} by solving the deformation problem with iterative matching, as exemplified by the non-rigid ICP method~\cite{amberg2007cvpr}.
More recently, some work also exploited deep learning for this purpose~\cite{liu2019cvpr, gu2019cvpr}.
These methods perform well for small deformations. For larger deformations, we need to handle the matching step explicitly.
A common approach is to use deformation graphs~\cite{sumner2007embedded}, 
where node-wise transformations are optimized based on precomputed correspondences.

In agricultural settings, correspondence estimation is particularly difficult due to repetitive structures, organic deformations, plant growth, and missing or occluded parts.
Several approaches addressed this by exploiting skeletal representations of plants~\cite{chebrolu2020icra, magistri2020iros}. 
Heiwolt~\etalcite{heiwolt2023icra}, instead, proposed encoding leaf shape for plant organs recognition across different growing stages.
However, many existing methods rely on accurate and dense 3D point clouds to work effectively~\cite{xiang2022icicml, dong2017icra}.

More recently, non-rigid reconstruction has been studied in the context of real-world agricultural robotics.
Lobefaro~\etalcite{lobefaro2024spatio} proposed to use deformation graph-based methods to generate 4D plant maps from noisy RGB-D data, exploiting temporal matches in images~\cite{lobefaro2023iros}.
Subsequent works further improve temporal matching in agricultural scenarios by incorporating semantics. 
Lobefaro~\etalcite{lobefaro2025spatio} introduced a spatio-temporally consistent semantic mapping framework for fruit growth monitoring, where instance-segmented RGB-D observations are associated over time and integrated into a consistent 3D semantic map. 
Freeman and Kantor~\cite{freeman2025iros} proposed a transformer-based method to associate apple fruitlets across days and viewpoints, leveraging shape and spatial geometry cues to improve matching robustness.

These works primarily address reconstruction, alignment, and temporal association of plant observations.
In contrast, we use non-rigid registration as a means to obtain an actionable geometric approximation by transferring a previous-session reconstruction to the current session, which can directly guide view planning during data acquisition.
To make this transfer reliable in practice, we adopt different warm-start strategies for individual plants and greenhouse crop rows.


\begin{figure*}[!t]
\centering
\includegraphics[width=1.0\textwidth]{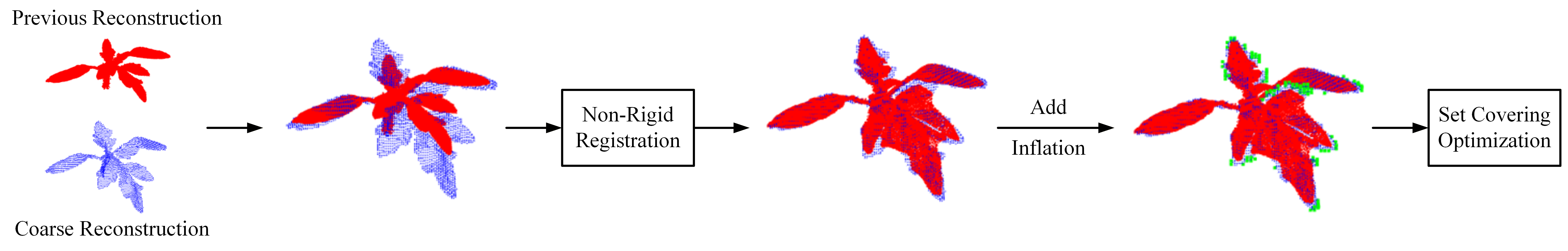}
\caption{
Overview of view planning guided by a previous-session reconstruction.
The individual-plant example uses observations from two consecutive monitoring sessions, illustrating growth-related changes in leaf extent and configuration.
The previous-session reconstruction is shown in red, while the coarse current-session reconstruction is shown in blue.
Non-rigid registration aligns the previous reconstruction with the current observations to obtain a transferred geometric approximation.
It then applies a conservative inflation (green) to account for inter-session changes and residual registration errors.
Finally, a set-cover optimization is solved to select the minimum view set that achieves the desired surface coverage of the inflated planning geometry.
}
\label{fig_view_planning}
\end{figure*}

\section{Our Approach}

We build on established techniques for non-rigid registration, set-cover optimization, and TSP-based routing.
We integrate these components into a one-shot view-planning pipeline~\cite{pan2024tro}, which selects an informative set of views and then optimizes their execution order, as shown in Fig.~\ref{fig_pipeline}.
Our task-specific design makes a previous-session reconstruction actionable for current-session planning through warm-started reconstruction transfer, conservative inflation, and setting-dependent view parameterization.
The following sections define the planning problem, describe reconstruction transfer and view selection, and present global path planning.

\subsection{Problem Definition}
We study view planning for \emph{repeated plant monitoring} guided by a previous-session reconstruction.
Given a plant 3D model built from the previous monitoring session and a coarse 3D reconstruction in the current session, the goal is to plan a sequence of sensor views that efficiently reconstructs the plant.

\subsubsection{Inputs}
Let $\mathcal{P}$ denote the previous-session reconstruction and let $\mathcal{Q}$ denote a coarse reconstruction of the current session (obtained from a passive mapping warm-start or a short NBV warm-up).
Let $\mathcal{V}$ be the feasible discrete view space under the platform's constraints.
Each view $v\in\mathcal{V}$ is parameterized by a 3D camera position $s\in\mathbb{R}^3$ and a viewing direction represented by a 3D look-at target $\ell\in\mathbb{R}^3$ (we fix or ignore roll around the optical axis, which has negligible effect on visibility~\cite{pan2024tro,zaenker2021viewpoint}).
Thus, the view parameterization involves two coupled decisions: \emph{where to stand} (camera position) and \emph{where to look} (look-at target).

\subsubsection{Outputs}
The planner outputs (i) a set of views $\mathcal{S}\subseteq\mathcal{V}$ to be executed in the current session, and (ii) an executable global path $\pi$ that orders these views for execution.

\subsubsection{Objectives}
The desired plan should (i) achieve high reconstruction quality by maximizing plant surface coverage, and (ii) minimize acquisition effort, measured by the number of executed views and the robot path length required to visit them.
In repeated monitoring, the previous-session reconstruction $\mathcal{P}$ provides plant-specific geometric information that can be transferred to the current session and exploited to improve both coverage and efficiency compared with planning from scratch.

\subsection{View Planning Guided by Previous-Session Reconstruction}
One-shot set-cover planning evaluates coverage on a target surface; hence it requires an explicit geometric proxy of the current plant surface.
We transfer the previous-session reconstruction to the current session via non-rigid registration and a conservative inflation step, yielding an actionable geometric approximation for visibility reasoning and set-cover-based view selection.

The reconstruction-transfer and view-selection process is illustrated in Fig.~\ref{fig_view_planning} and consists of three core steps:
(1) \emph{non-rigid registration} to align the previous-session reconstruction with the current-session coarse reconstruction under plant deformation;
(2) \emph{conservative inflation} to obtain a robust planning geometry that accounts for inter-session changes and residual registration errors; and
(3) \emph{set-cover-based view selection} to choose a minimal set of views that covers the inflated geometry approximation.

\subsubsection{Non-Rigid Registration}
We assume that a coarse cross-session alignment is available, so the previous-session reconstruction and the current-session coarse reconstruction are expressed in a common coordinate system.
Let the previous-session source point cloud be $\mathcal{P}=\{p_i\}_{i=1}^{N}$ and the current-session coarse reconstruction be $\mathcal{Q}$.
We seek a non-rigid warp that transfers the previous geometry to the current session for subsequent visibility reasoning and view selection.

Following deformation-graph approaches~\cite{sumner2007embedded,lobefaro2024spatio}, we obtain sparse control nodes $\{n_j\}_{j=1}^{M}$ by voxel downsampling of $\mathcal{P}$ at resolution $r_{\mathrm{graph}}$ and connect spatially neighboring nodes.
Each node carries a local rigid transformation parameterized by a rotation $R_j\in\mathrm{SO}(3)$ and a translation $t_j\in\mathbb{R}^{3}$.
Each source point is associated with nearby control nodes using normalized Gaussian distance weights.
The node transformations collectively define a spatially varying deformation field:
\begin{equation}
\tilde p_i = F_{\Theta}(p_i),
\qquad
\Theta=\{R_j,t_j\}_{j=1}^{M}.
\end{equation}
This representation accommodates local changes in leaf positions and orientations without requiring the entire plant to undergo a single rigid motion.

We estimate the deformation by combining data alignment with spatial regularization:
\begin{equation}
\label{eq:nricp_obj}
\mathcal{L}
=
\lambda_{\mathrm{data}}\mathcal{L}_{\mathrm{data}}
+
\lambda_{\mathrm{reg}}\mathcal{L}_{\mathrm{reg}}.
\end{equation}
The data term measures geometric discrepancies between the deformed source and the current observations.
The graph regularizer discourages spatially inconsistent local transformations, coupling neighboring control nodes and constraining deformation in regions with limited observations.
The current coarse reconstruction therefore provides alignment evidence, while the graph regularization controls how this evidence influences the transferred geometry.
We instantiate the deformation interpolation, alignment term, and regularizer according to the reconstruction setting, with implementation details provided in the experimental sections.
The resulting aligned point cloud $\tilde{\mathcal{P}}=\{\tilde p_i\}$ is passed to the inflation and view-selection stages.

\subsubsection{Conservative Inflation}
Inter-session changes and residual registration errors can leave currently observed structures insufficiently represented in the aligned previous-session reconstruction.
We therefore augment the transferred geometry with a conservative inflation set $\mathcal{I}$.

We consider candidate locations on a regular three-dimensional grid within the bounded plant region.
The inflation criterion retains locations whose nearest-neighbor distance to the current coarse reconstruction $\mathcal{Q}$ is below $\gamma_{\mathrm{near}}$ and whose distance to the aligned reconstruction $\tilde{\mathcal{P}}$ exceeds $\gamma_{\mathrm{far}}$.
This adds a spatial neighborhood around current observations where the transferred geometry lacks nearby support.
The grid spacing and the setting-specific implementations of candidate generation and filtering are detailed in the experimental sections.

We also retain $\mathcal{Q}$ as an occlusion surface so that already observed structures are considered during subsequent ray casting.
The resulting planning approximation is
\begin{equation}
\tilde{\mathcal{P}}^+
=
\tilde{\mathcal{P}}\cup\mathcal{Q}\cup\mathcal{I}.
\end{equation}

\begin{figure}[!t]
\centering
\includegraphics[width=0.8\columnwidth]{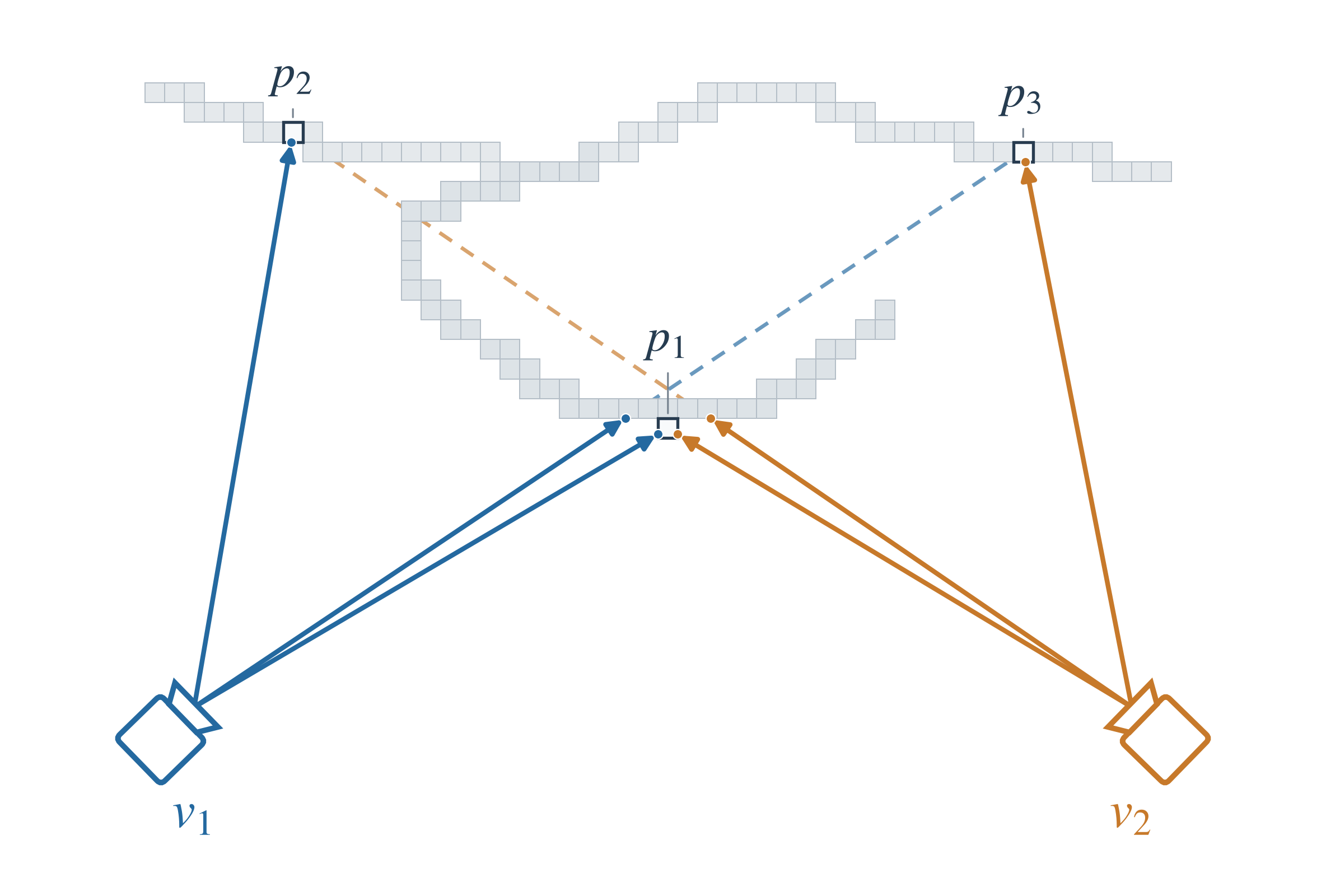}
\caption{
First-hit visibility evaluation for two candidate views.
Gray cells represent the voxelized planning geometry, with three representative surface voxels labeled $p_1$--$p_3$.
Solid colored arrows terminate at the first occupied voxels encountered by the rays; dashed continuations indicate occluded regions beyond these hits.
Both views observe $p_1$, whereas $p_2$ is visible only from $v_1$ and $p_3$ only from $v_2$, illustrating shared coverage and view-dependent occlusion.
}
\label{fig:visibility_indicator}
\end{figure}

\subsubsection{Set-Cover-Based View Selection}
To make optimization tractable, we represent the planning geometry $\tilde{\mathcal{P}}^{+}$ using OctoMap~\citep{hornung2013ar} at resolution $r_{\mathrm{cover}}$.
Each occupied voxel is represented by its center point, yielding the discretized surface set $\mathcal{P}_{\mathit{surf}}=\{p_i\}$.

For each candidate view $v$, we predict the surface voxels observable by the camera using projection and first-hit ray casting.
We first project occupied-voxel centers into the image using the candidate camera pose and camera intrinsics, discarding projections behind the camera or outside the image.
For each valid projection, we cast a ray from the camera through the corresponding image pixel.
The first occupied voxel encountered within the sensing range is recorded as an observable surface voxel, accounting for occlusion by intervening geometry.
Collecting and deduplicating these first-hit voxels gives the visible subset $\mathcal{P}_v \subseteq \mathcal{P}_{\mathit{surf}}$.
We then define the binary visibility indicator
\begin{equation}
\label{equ:indicator}
I(p,v) =
\begin{cases}
1, & \text{if } p \in \mathcal{P}_v,\\
0, & \text{otherwise.}
\end{cases}
\end{equation}
Physically, $I(p,v)=1$ means that view $v$ is predicted to observe the surface voxel represented by $p$ on the planning geometry, while $I(p,v)=0$ means that this voxel is not included in the predicted observation.
These binary visibility relations specify which candidate views can satisfy each surface-coverage constraint in the subsequent set-cover problem.
Fig.~\ref{fig:visibility_indicator} illustrates the first-hit visibility test on a voxelized surface.
Both candidate views observe $p_1$, giving $I(p_1,v_1)=I(p_1,v_2)=1$.
The foreground fold occludes $p_2$ from $v_2$ and $p_3$ from $v_1$, while these voxels remain visible from $v_1$ and $v_2$, respectively.
The resulting visibility sets therefore contain both shared and complementary surface coverage, which the set-cover formulation uses to select views jointly.

Already executed views, including warm-start views, are excluded from the decision set $\mathcal{V}$.
Let $\mathcal{P}_{\mathrm{pre}}$ denote the planning-surface voxels already covered by the warm-start observations.
We restrict the coverage demand to voxels observable from at least one remaining candidate view and not already covered:
\begin{equation}
\mathcal{D}
=
\left(\bigcup_{v\in\mathcal{V}}\mathcal{P}_v\right)
\setminus\mathcal{P}_{\mathrm{pre}}.
\end{equation}
This restriction avoids imposing coverage constraints on geometry that cannot be observed from the available candidate set.

We then select a minimum-cardinality view set that covers this demand:
\begin{equation}
\label{eq:setcover}
\begin{aligned}
\min_{\{x_v\}} \quad & \sum_{v\in\mathcal{V}}x_v \\
\text{s.t.}\quad
& \sum_{v\in\mathcal{V}}I(p,v)x_v \ge 1,
&& \forall\,p\in\mathcal{D}, \\
& x_v\in\{0,1\},
&& \forall\,v\in\mathcal{V}.
\end{aligned}
\end{equation}
Here, $x_v$ indicates whether candidate view $v$ is selected.
Each demanded surface voxel must be observed by at least one selected view.
We solve Eq.~\eqref{eq:setcover} with Gurobi~\citep{gurobi2021gurobi}.
The selected view set is
\begin{equation}
\mathcal{S}
=
\{v\in\mathcal{V}\mid x_v=1\},
\end{equation}
which is passed to the path planning module to compute the global execution path $\pi$.

\subsection{Global Path Planning}
Given the view set $\mathcal{S}$ selected in the preceding stage and a starting pose $v_0$, we compute an executable global path $\pi = (v_0, v_1, \ldots, v_{|\mathcal{S}|})$ that visits each selected view exactly once.
Let $d(v_i,v_j)$ denote the motion-cost estimate used to order two views, with setting-specific definitions given in the experimental sections.
We minimize the total travel cost:
\begin{equation}
\label{eq:global_path}
\min_{\pi}\;\sum_{i=0}^{|\mathcal{S}|-1} d\!\left(v_i, v_{i+1}\right),
\quad \text{s.t.}\;\{v_1,\ldots,v_{|\mathcal{S}|}\}=\mathcal{S}.
\end{equation}
This is the shortest Hamiltonian path problem, i.e., an open traveling-salesman problem (no return to the start)~\cite{osswald2016ral}.
We solve Eq.~\eqref{eq:global_path} to optimality as an open-TSP problem with a fixed start and, when required by the platform, a fixed endpoint.
The setting-specific solvers are described in the experimental sections.
The resulting global path $\pi$ is then executed by the robot to acquire the selected views.

\section{Simulation-Based Evaluation Using Real-World Plant Data}
\label{sec:exp}

We organize experiments to directly validate our three contributions across two complementary monitoring settings.
The individual-plant setting provides a controlled benchmark to evaluate our one-shot view planning and to ablate the previous-session reconstruction transfer pipeline, including warm-started non-rigid registration and conservative inflation.
The greenhouse-row setting evaluates the approach at a larger scale under restricted fields of view, where we study the setting-dependent view parameterization and the effect of transferred geometry on 3D look-at planning.
In addition, we conduct a pose-noise robustness study to assess sensitivity to execution errors and whether lightweight registration refinement can recover performance.

\subsection{Experimental Setup}

\subsubsection{Test Platform}
We adopt the view-planning simulator from~\cite{pan2024tro}, which implements a virtual sensor via OctoMap-based ray casting: given a camera pose specified by a 3D position and a 3D look-at target, it renders a point cloud of visible surface points from this viewpoint.
This design is well-suited to our datasets, where the ground-truth reference models are derived from real measurements and may be noisy and non-watertight. 
Ray casting on an occupancy representation avoids requiring a perfectly watertight mesh.
All virtual images are rendered at $1280 \times 720$ resolution, using camera intrinsics copied from an Intel RealSense D435.
All simulation-based experiments are executed on a PC with an Intel Core i7-12700H CPU (14 cores) and 32\,GB RAM.
We assume a stop-and-sense acquisition model: measurements are acquired only at discrete planned views (as in practice, pose stabilization and exposure need to be considered).
We do not integrate measurements collected while moving to ensure fair comparison across methods.

\subsubsection{Evaluation Metrics}
We evaluate methods along three axes: (i) planning effectiveness in terms of surface coverage, (ii) acquisition cost in terms of sensing and motion resources, and (iii) robustness in terms of geometric fidelity under pose perturbations.

\textbf{Surface Coverage.}
Since our simulator renders noise-free measurements, a coverage-based metric provides the most direct and reliable measure of reconstruction performance.
However, real plant models in our datasets can be non-watertight and may contain inner surfaces, and in the greenhouse-row setting the sensor is physically constrained to observe the crop mainly from one side.
Therefore, a non-trivial fraction of surface points is fundamentally unobservable from the feasible view space; including such points in the denominator would unfairly penalize all methods and confound performance comparisons.

Following~\cite{pan2024tro}, we normalize coverage by an oracle visible set computed on the ground-truth reference geometry.
Let $\mathcal{G}_{\mathit{surf}}$ denote the voxelized reference surface and let $\mathcal{G}_v\subseteq\mathcal{G}_{\mathit{surf}}$ denote the surface voxels visible from view $v$ under the virtual sensor.
These evaluation sets are computed on the reference geometry independently of the transferred planning geometry.

Let $\mathcal{V}_{\mathrm{ref}}$ denote the reference view space used for oracle construction, which is fixed across the compared methods for each evaluation scene.
The oracle-visible surface set is
\begin{equation}
\label{eq:Omega}
\Omega
=
\bigcup_{v\in\mathcal{V}_{\mathrm{ref}}}\mathcal{G}_v.
\end{equation}
Let $\mathcal{S}_{\mathrm{exec}}$ denote all views acquired up to the current step, including warm-start observations.
The accumulated observed surface is
\begin{equation}
C
=
\bigcup_{v\in\mathcal{S}_{\mathrm{exec}}}\mathcal{G}_v.
\end{equation}
We report oracle-visible surface coverage as
\begin{equation}
\frac{|C|}{|\Omega|}.
\end{equation}

\textbf{Acquisition Cost: Number of Views and Path Length.}
To quantify sensing and motion resources, we report:
(a) the number of executed views, and (b) the travel cost measured as the accumulated Euclidean path length connecting consecutive executed views.
We include the number of views since each executed view typically incurs a non-negligible stop-and-sense overhead (sensing and processing) and provides a hardware-agnostic measure of sampling effort.
For motion cost, one-shot methods compute the execution order using our global path planning module after view selection, whereas sequential NBV baselines accumulate path length in their native execution order (i.e., without reordering), reflecting their online nature.

\textbf{Robustness: Chamfer Distance and F-Score.}
Coverage is most informative in the noise-free simulator, but does not capture geometric distortions caused by execution errors.
To evaluate robustness, we conduct a pose-noise study and compare the final reconstruction against the ground-truth surface point cloud (as a geometric fidelity metric, independent of the oracle-visible set used for coverage).
We report: (a) symmetric Chamfer Distance, and (b) F-score at two distance thresholds \(\tau \in \{0.01, 0.02\}\)~m, including precision, recall, and \(F_1\).
These geometry metrics jointly reflect surface accuracy and completeness under pose perturbations and remain meaningful even when parts of the plant are unobservable.

The above setup and metrics are shared across both settings; we next describe setting-specific setups in their own sections.

\subsection{Individual-Plant Evaluation}

In this section, we evaluate our one-shot view planning on individual plants and ablate the previous-session reconstruction transfer pipeline (warm-started registration and inflation).

\subsubsection{Data and Protocol}
We use real multi-session scans of growing individual plants from~\cite{chebrolu2021registration}, which include maize and tomato, to evaluate previous-session reconstruction transfer and view planning under non-rigid inter-session deformation.
The dataset contains two plant types, each with three plant instances.
For each instance, we use the last two consecutive monitoring sessions, as later growth stages exhibit more complex geometry and occlusion and thus provide a more challenging testbed for view planning.
Since view planning algorithms are often sensitive to the initial observation, we evaluate two global rotations about the $Z^+$ axis (0\degree and 45\degree) applied to the plant model and three initial viewpoints (near-horizontal, oblique, and near-top), following~\cite{pan2024tro}.
For each view-space configuration, this yields $2$ (plant types) $\times\,3$ (instances) $\times\,2$ (rotations) $\times\,3$ (initial views) $=36$ test cases.

\begin{figure}[!t]
\centering
\includegraphics[width=1.0\columnwidth]{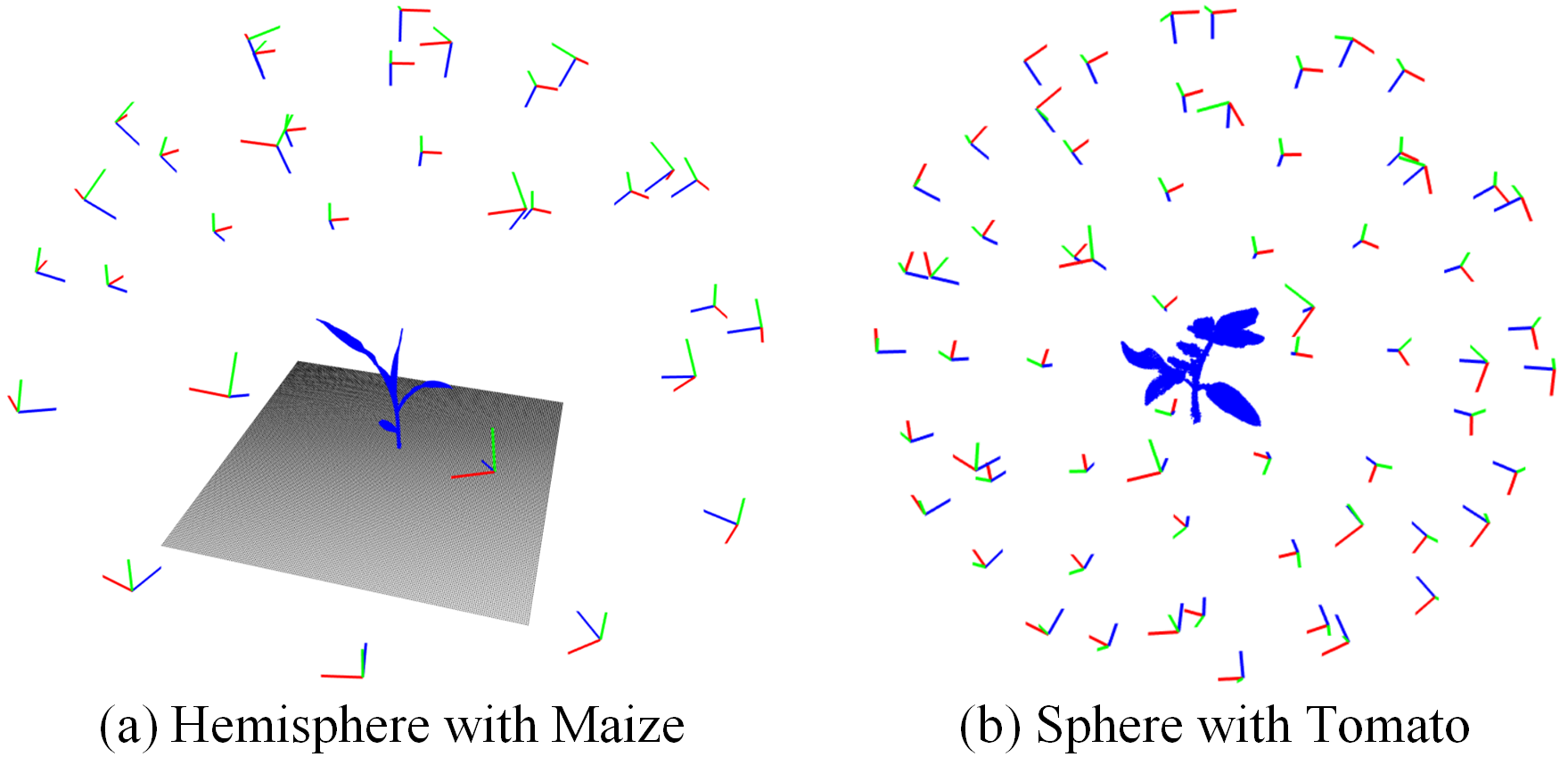}
\caption{
Examples of the two environmental configurations used in our experiments.
(a) Hemisphere view space with a supporting table, shown with a Maize plant.
(b) Sphere view space without a table, shown with a Tomato plant.
}
\label{fig_view_config}
\vspace{-0.2cm}
\end{figure}

\subsubsection{View Space Configuration and Parameterization}
We follow the established object-centric reconstruction benchmark in~\cite{pan2024tro}, using a hemispherical view space of 32 candidate camera positions to maintain comparability with existing methods.
To evaluate planning beyond the tabletop configuration, we additionally consider a spherical view space of 63 candidate positions, allowing observations from below the plant.
These numbers specify the discretized candidate spaces rather than the numbers of views selected for acquisition.
For each configuration, we optimize the spatial distribution of the prescribed number of candidate positions by solving the Tammes problem~\citep{lai2023iterated}, which maximizes the minimum pairwise distance between positions on the spherical surface.
This produces approximately uniform spatial sampling without favoring particular viewing regions.
Fig.~\ref{fig_view_config} illustrates the hemispherical~(32) and spherical~(63) configurations.

In the individual-plant setting, we discretize only the 3D camera positions using the candidate sets above.
The camera orientation is constrained by a fixed look-at direction toward the plant centroid, implemented by pointing the optical axis to the center of the plant's bounding box.
Prior work suggests that object-centric planning is generally insensitive to small perturbations of the look-at target~\cite{pan2024tro}; for simplicity, we therefore use the bounding-box center computed from the reference model as the look-at target.
As a result, the nominal 6D view planning problem reduces to 3D position selection.
In this setting, the oracle visible set $\Omega$ in Eq.~(\ref{eq:Omega}) is obtained by taking the union of the visible sets over all hemispherical/spherical candidate views.
The OctoMap resolution used by the virtual sensor (i.e., the voxel size for coverage evaluation on the reference model) is set to 0.002 m to capture fine surface details.
We normalize the overall plant size to 0.12\,m and place candidate viewpoints on a hemisphere/sphere of radius 0.4\,m centered at the plant.

\begin{figure}[!t]
\centering
\includegraphics[width=0.85\columnwidth]{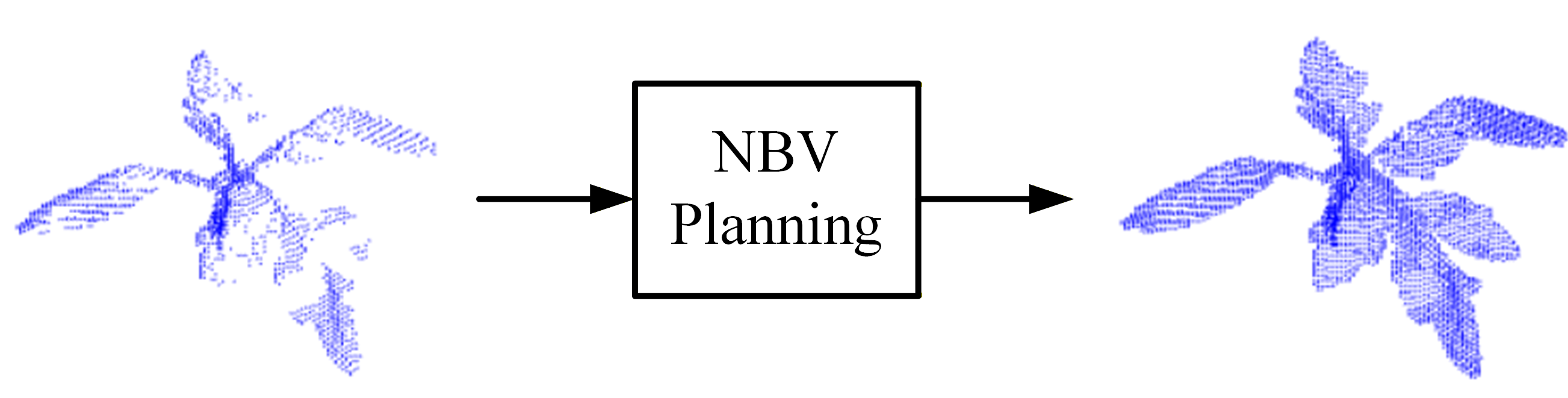}
\caption{
One-step NBV warm-up for coarse reconstruction.
Starting from an initial partial scan (left), we run a brief NBV stage (center) to select one additional informative view and fuse its observation, producing a more complete coarse reconstruction (right).
This warm-start reconstruction is used as the target for non-rigid registration.
}
\label{fig_corase_nbv}
\vspace{-0.2cm}
\end{figure}

\subsubsection{Setting-Specific Implementation}
To obtain a warm start for non-rigid registration, we adopt the short NBV warm-up strategy in~\cite{pan2024tro} to improve the quality of an initial partial reconstruction, and use it here to initialize cross-session non-rigid alignment.
Concretely, we run a brief NBV stage with PC-NBV~\cite{zeng2020iros} to select one additional view; the resulting observation is fused into the initial scan to form a coarse reconstruction, which is then used as the target for our non-rigid registration module.
As illustrated in Fig.~\ref{fig_corase_nbv}, starting from an initial partial scan, the NBV module selects one additional informative view whose observation is fused into the reconstruction.

For non-rigid registration, we use the standard embedded-deformation model~\cite{sumner2007embedded} with control-node downsampling resolution $r_{\mathrm{graph}}=0.004$\,m.
Each source point is associated with its four nearest control nodes.
The deformation is computed by blending the positions obtained under their local rigid transformations:
\begin{equation}
F_{\Theta}(p_i)
=
\sum_{n_j\in\mathcal{A}(i)}
w_{ij}
\Bigl[R_j(p_i-n_j)+n_j+t_j\Bigr],
\end{equation}
where $\mathcal{A}(i)$ is the anchor set of $p_i$ and the Gaussian distance weights satisfy $\sum_{n_j\in\mathcal{A}(i)}w_{ij}=1$.
The data term in Eq.~\eqref{eq:nricp_obj} is a sampled, truncated symmetric Chamfer loss between the deformed source and the current coarse reconstruction.
At each iteration, we randomly sample up to 1,000 points from each cloud to evaluate this term.
The graph regularizer is the ED as-rigid-as-possible (ARAP) penalty, which penalizes discrepancies between a neighboring node's position predicted by a local rigid transformation and its position under its own transformation.
We set $\lambda_{\mathrm{data}}=0.1$ and $\lambda_{\mathrm{reg}}=1.0$.
Node rotations and translations are optimized using Adam~\cite{kingma2014adam} for up to 300 iterations with an initial learning rate of 0.001.

For inflation, we traverse a regular three-dimensional grid within the plant bounding region at a spacing of 0.004\,m.
For each grid location, two KD-tree queries obtain its nearest-neighbor distances to the current coarse reconstruction and the aligned previous-session reconstruction.
We retain locations satisfying the near-current and far-history conditions with $\gamma_{\mathrm{near}}=0.003$\,m and $\gamma_{\mathrm{far}}=0.005$\,m, respectively.

For set covering, we voxelize the inflated approximation at resolution $r_{\mathrm{cover}}=0.004$\,m when constructing $\mathcal{P}_{\mathit{surf}}$ and visibility sets.

For path planning, we compute the collision-free local path length using a simple collision model that approximates the plant as a bounding sphere, following~\cite{pan2024tro}.
The resulting open TSP is solved by exact subset dynamic programming~\cite{held1962dynamic}.

\subsubsection{Baselines}
We compare against two strong baselines reported in~\cite{pan2024tro}: GMC~\cite{pan2023cviu} and MA-SCVP~\cite{pan2024tro}.
GMC is an NBV-based approach that iteratively selects the next view by maximizing an information-gain objective.
MA-SCVP is a combined pipeline that triggers a set-covering stage after a short NBV warm-up, using a learned shape prior to guide view selection.

\begin{table}[!t]
  \centering
  \renewcommand{\arraystretch}{1.15}
  \resizebox{\columnwidth}{!}{%
  \begin{tabular}{|c|c|c|c|c|}
    \hline
    \makecell[c]{View Space} & Method & \makecell[c]{Number\\of Views} & 
    \makecell[c]{Surface\\Coverage (\%)} & 
    \makecell[c]{Path\\Length (m)} \\ \Xhline{1.2pt}
    \multirow{6}{*}{Hemisphere}
      & \multirow{4}{*}{GMC~\cite{pan2023cviu}} & 5  & 91.12 $\pm$ 3.12 & \textbf{2.05} $\pm$ 0.38 \\ \cline{3-5}
      &                      & 10 & 97.71 $\pm$ 0.83 & 4.54 $\pm$ 0.67 \\ \cline{3-5}
      &                      & 15 & 98.67 $\pm$ 0.50 & 6.02 $\pm$ 0.97 \\ \cline{3-5}
      &                      & 20 & \textbf{99.11} $\pm$ 0.53 & 7.17 $\pm$ 1.04 \\ \cline{2-5}
      & MA-SCVP~\cite{pan2024tro}             & 12.61 $\pm$ 1.20 & 98.80 $\pm$ 0.43 & 3.05 $\pm$ 0.30 \\ \cline{2-5}
      & Ours          & 11.06 $\pm$ 3.38 & 98.77 $\pm$ 0.91 & 3.10 $\pm$ 0.51 \\ \Xhline{1.2pt}
    \multirow{6}{*}{Sphere}
      & \multirow{4}{*}{GMC~\cite{pan2023cviu}} & 5  & 81.86 $\pm$ 11.13 & \textbf{1.77} $\pm$ 0.51 \\ \cline{3-5}
      &                      & 10 & 90.58 $\pm$ 10.25 & 4.21 $\pm$ 0.82 \\ \cline{3-5}
      &                      & 15 & 95.18 $\pm$ 5.84 & 6.75 $\pm$ 1.04 \\ \cline{3-5}
      &                      & 20 & 97.14 $\pm$ 3.63 & 9.21 $\pm$ 1.26 \\ \cline{2-5}
      & MA-SCVP~\cite{pan2024tro}             & 12.72 $\pm$ 1.09 & 92.92 $\pm$ 4.29 & 2.99 $\pm$ 0.22 \\ \cline{2-5}
      & Ours          & 9.22 $\pm$ 1.66 & \textbf{98.44} $\pm$ 0.68 & 3.35 $\pm$ 0.41 \\ \Xhline{1.2pt}
  \end{tabular}
  }%
 \caption{
Coverage--effort trade-off under different view-space configurations (hemisphere vs.\ sphere).
GMC selects views sequentially (NBV), hence results are shown for fixed budgets of 5/10/15/20 views, illustrating the coverage gain versus increasing motion and sensing effort.
In contrast, MA-SCVP and our method perform one-shot view-set selection (set cover), producing a variable-sized view set per test case.
Overall, our method achieves comparable coverage to MA-SCVP on the hemisphere while using a similar number of views, and it substantially improves coverage on the sphere by exploiting additional feasible viewpoints (including bottom views), with fewer executed views than MA-SCVP.
}
  \label{tab:evaluation}
  \vspace{-0.2cm}
\end{table}

\subsubsection{Evaluation of View Planning Performance}

Table~\ref{tab:evaluation} compares the coverage--effort trade-off (coverage vs.\ executed views and path length) under two view-space configurations.

\textbf{Comparison to NBV-based GMC.} GMC increases coverage as the view budget grows, but at the cost of substantially longer execution paths. For example, on the hemisphere, coverage improves from 91.12\% (5 views) to 99.11\% (20 views), while the path length increases from 2.05\,m to 7.17\,m.
In contrast, our one-shot planner achieves 98.77\% coverage with 11.06 views and a 3.10\,m path, offering a more favorable coverage--effort trade-off.

\begin{figure}[!t]
\centering
\includegraphics[width=1.0\columnwidth]{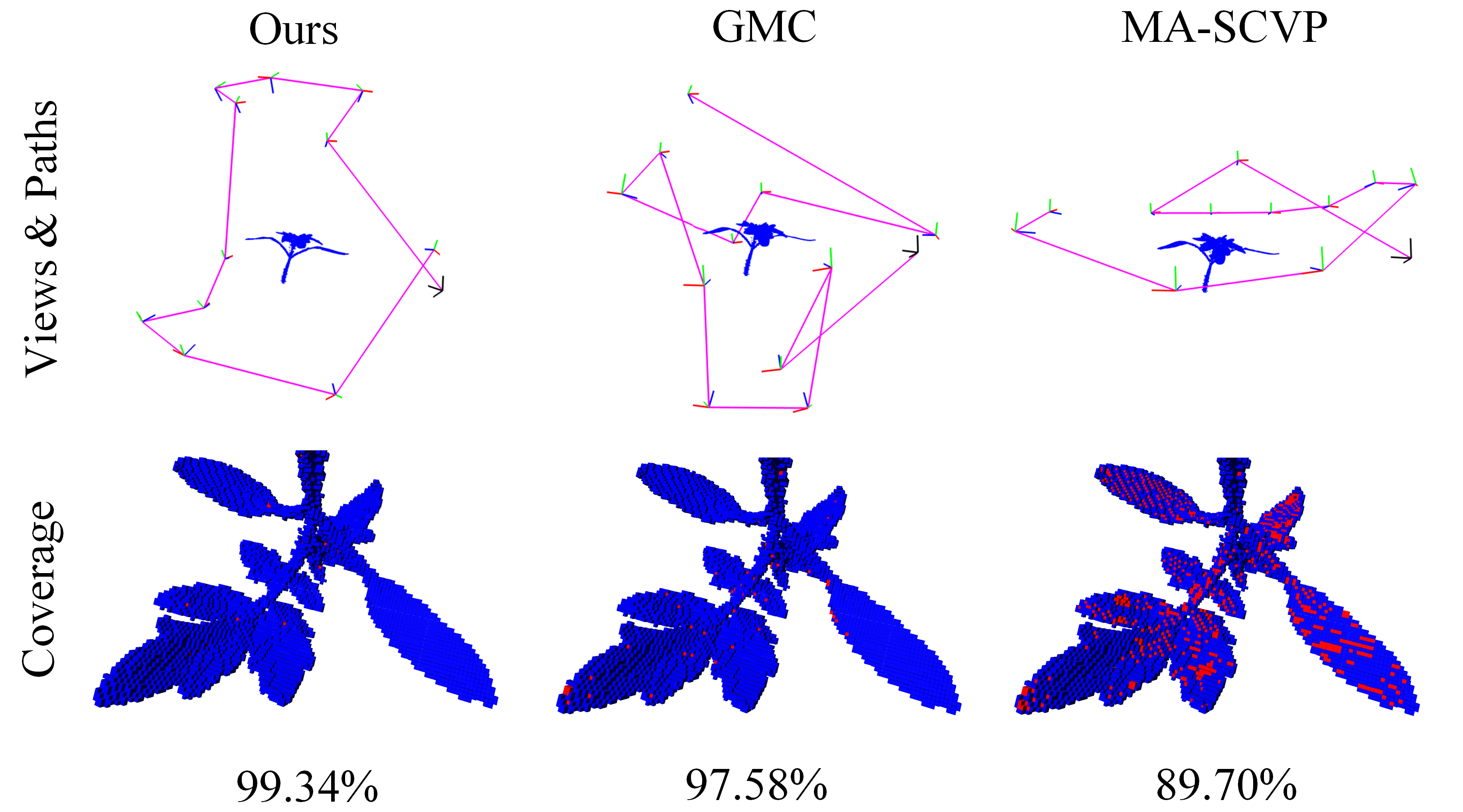}
\caption{
Visual comparison of our method with GMC~\cite{pan2023cviu} and \mbox{MA-SCVP~\cite{pan2024tro}}.
The top row shows the same initial view (black), the selected views (red-green-blue), and the corresponding view paths~(purple).
For a fair comparison, all methods are visualized with 12~views (our method and \mbox{MA-SCVP} both predict 12 views, while 12 views are selected from GMC for consistency).
The bottom row visualizes the reconstructed plant surfaces (blue) and uncovered regions (red) using voxelization, focusing on the bottom part of the plant.
Our method achieves the highest surface coverage with a shorter path; GMC also attains good coverage but with a longer path; MA-SCVP, despite a similar path length, fails to observe the bottom surface details.
}
\label{fig_compare_sphere}
\end{figure}

\textbf{Comparison to MA-SCVP.}
MA-SCVP uses more views on average in the hemispherical setting (12.61 versus 11.06), which may provide additional observations of small, partially occluded surface regions.
This offers a plausible explanation for its marginally higher coverage (98.80\% versus 98.77\%), while our method achieves comparable coverage with fewer acquired views.
On the sphere, our method achieves higher coverage (98.44\% vs.\ 92.92\%) while using fewer executed views on average (9.22 vs.\ 12.72).
This gain comes from exploiting the expanded feasible view space (including bottom views), which MA-SCVP cannot represent due to its fixed 32-pose hemisphere parameterization.
Fig.~\ref{fig_compare_sphere} visualizes a representative sphere-case example.

\begin{table}[!t]
  \centering
  \renewcommand{\arraystretch}{1.15}
  \resizebox{0.8\columnwidth}{!}{%
  \begin{tabular}{|c|c|c|c|}
    \hline
    \makecell[c]{Plant Type} & \makecell[c]{View Space} & 
    \makecell[c]{Number\\of Views} & 
    \makecell[c]{Surface\\Coverage (\%)} \\ \Xhline{1.2pt}

    \multirow{2}{*}{Maize}
      & Hemisphere & 7.94 $\pm$ 1.11 & 98.09 $\pm$ 0.82 \\ \cline{2-4}
      & Sphere     & \textbf{7.89} $\pm$ 0.96 & 98.02 $\pm$ 0.69 \\ \Xhline{1.2pt}

    \multirow{2}{*}{Tomato}
      & Hemisphere & 14.17 $\pm$ 1.34 & 99.45 $\pm$ 0.20 \\ \cline{2-4}
      & Sphere     & \textbf{10.56} $\pm$ 0.98 & 98.85 $\pm$ 0.32  \\ \Xhline{1.2pt}
  \end{tabular}
  }
\caption{
Performance of our method across plant types and view-space configurations.
The method allocates more views to the structurally more occluded tomato than to maize, and it requires fewer views for tomato when bottom viewpoints are available (sphere vs.\ hemisphere) while maintaining comparable coverage.
}
  \label{tab:complexity}
  \vspace{-0.2cm}
\end{table}

\textbf{Planning-Time Analysis.}
We compare the computation time of different planners for generating view sets and execution orders.
In our setup, planning takes 10--15\,s for our method, less than 1\,s for MA-SCVP, and about 1.5\,s per iteration for GMC (about 30\,s for 20 iterations/views).
While our planner is slower than MA-SCVP, it provides improved coverage and supports flexible view-space configurations (e.g., the sphere setting).

\subsubsection{Analysis on Plant Complexity}

We evaluate our method across different view space configurations and plant types, which correspond to different levels of structural complexity.
As shown in Table~\ref{tab:complexity}, our method assigns fewer views to the simpler plant (maize) and more to the more complex plant~(tomato), consistent with their occlusion patterns; tomato’s broad leaves incur heavier self-occlusion.
Moreover, for the tomato, we require fewer views in the sphere configuration than in the hemisphere configuration, because bottom occlusions are easier to resolve when viewpoints are allowed below the plant.
Taken together, these results indicate that our method adaptively allocates the viewpoint budget according to plant complexity and the available view space.

\subsubsection{Ablation Study on Inflation}

We present an ablation study on the impact of the inflation module, where only the aligned previous-session reconstruction is retained for the set covering.
As shown in~Table~\ref{tab:ablation_inflation}, incorporating inflation consistently improves surface coverage compared to the no-inflation variant across both plant types.
The improvement is particularly notable for maize, which we attribute to inflation compensating for residual alignment errors and capturing geometry that appears in the current coarse reconstruction but is missing from the aligned previous-session reconstruction.

\subsubsection{Analysis on NBV Module}

We analyze the effect of the NBV module by computing the Chamfer distance between the aligned transferred reconstructions and the ground-truth surface.
As shown in Table~\ref{tab:registration}, introducing a single NBV consistently reduces the Chamfer distance compared with using only the initial view.
This indicates that without the NBV, the non-rigid registration is less reliable and more prone to errors.
Moreover, when relying solely on the initial view (without inflation), the final reconstruction surface coverage decreases to $97.61\,\pm\,1.75$, averaged over both the hemisphere and sphere.
These results demonstrate that adding one NBV provides complementary information that enhances the robustness of non-rigid registration and ultimately improves the reconstruction quality.

\begin{table}[!t]
  \centering
  \renewcommand{\arraystretch}{1.15}
  \resizebox{\columnwidth}{!}{%
  \begin{tabular}{|c|c|c|c|c|}
    \hline
    \makecell[c]{Plant Type} & Method & \makecell[c]{Number\\of Views} & 
    \makecell[c]{Surface\\Coverage (\%)} & 
    \makecell[c]{Path\\Length (m)} \\ \Xhline{1.2pt}

    \multirow{2}{*}{Maize}
      & Ours w/o Inflation & 6.75 $\pm$ 0.97 & 96.99 $\pm$ 1.16 & 2.51 $\pm$ 0.32 \\ \cline{2-5}
      & Ours w/ Inflation  & 7.92 $\pm$ 1.02 & \textbf{98.06} $\pm$ 0.75 & 2.85 $\pm$ 0.29 \\ \Xhline{1.2pt}

    \multirow{2}{*}{Tomato}
      & Ours w/o Inflation & 11.42 $\pm$ 2.05 & 98.98 $\pm$ 0.50 & 3.38 $\pm$ 0.30 \\ \cline{2-5}
      & Ours w/ Inflation  & 12.37 $\pm$ 2.17 & \textbf{99.15} $\pm$ 0.40 & 3.61 $\pm$ 0.28 \\ \Xhline{1.2pt}
  \end{tabular}
  }
\caption{
Ablation on conservative inflation for planning-geometry construction.
Inflation improves surface coverage for both plant types with only a small increase in the number of executed views and path length, indicating improved robustness to inter-session changes and residual registration errors.
}
  \label{tab:ablation_inflation}
\end{table}

\begin{table}[!t]
  \centering
  \renewcommand{\arraystretch}{1.15}
  \resizebox{0.75\columnwidth}{!}{%
  \begin{tabular}{|c|c|}
    \hline
    \makecell[c]{Method} & \makecell[c]{Chamfer Distance (mm)} \\ \Xhline{1.2pt}
     Initial View Only & 2.764 $\pm$ 1.195 \\  \hline
     Initial View + One NBV & \textbf{1.743} $\pm$ 0.444 \\ \Xhline{1.2pt}
  \end{tabular}
  }
\caption{
Registration quality for the warm-start stage, comparing alignment using only the initial view versus adding one NBV view.
Adding one NBV reduces the Chamfer distance, indicating a more reliable alignment target for transferring the previous-session reconstruction.
}
  \label{tab:registration}
\vspace{-0.2cm}
\end{table}

\subsubsection{Summary}
Overall, the individual-plant experiments lead to two main takeaways.
First, view planning guided by a previous-session reconstruction enables efficient one-shot planning under non-rigid inter-session changes, achieving high oracle-visible surface coverage with compact view sets and short execution paths across different view-space configurations (including cases not supported by fixed-parameterization MA-SCVP).
Second, the ablations highlight the importance of reliable reconstruction transfer: a short NBV warm start improves alignment quality, and the conservative inflation step consistently boosts coverage by compensating for residual registration errors and inter-session changes.

\subsection{Greenhouse-Row Evaluation}

In this section, we evaluate the proposed method on greenhouse crop rows under restricted fields of view, emphasizing setting-dependent view parameterization and 3D look-at planning guided by transferred geometry.

\begin{figure}[!t]
\centering
\includegraphics[width=1.0\columnwidth]{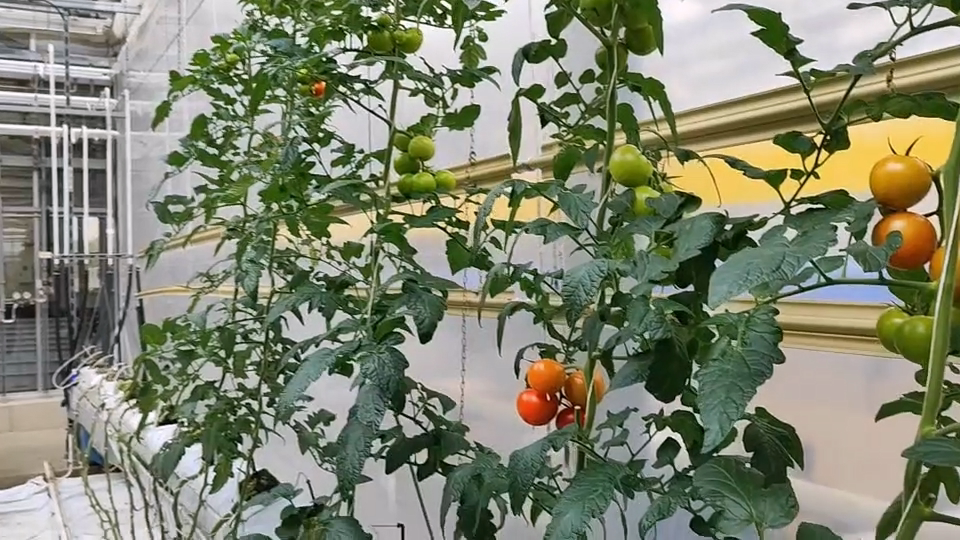}
\caption{
Greenhouse crop row used in our greenhouse-row dataset.
The row contains dense foliage and fruits, resulting in severe self-occlusion, which motivates repeated monitoring using view planning.
}
\label{fig_greenhouse_row}
\vspace{-0.2cm}
\end{figure}

\subsubsection{Data and Protocol}
We collected a greenhouse crop-row dataset containing two crop types (tomato and cherry tomato), captured over five monitoring sessions spanning two weeks (Sep.~15, 18, 22, 25, and 29, 2025).
In each session, we scanned one crop row per crop type.
Data were acquired using Polycam\footnote{\url{https://poly.cam/}} on an iPhone/iPad Pro equipped with a LiDAR sensor, which provides RGB images, per-frame depth maps, and estimated camera poses.
For each session, we reconstruct a reference 3D model using the Nerfstudio~\cite{tancik2023nerfstudio} depth-nerfacto pipeline and export a point cloud representation.
Fig.~\ref{fig_greenhouse_row} shows a representative greenhouse-row scene from our dataset.
We use this handheld acquisition to efficiently collect multi-session real-world data for simulation-based evaluation; the proposed planning method is agnostic to the sensing platform as long as a feasible view space can be specified.

To obtain clean reference geometry for simulation-based evaluation (ray casting and oracle visible-set construction), we apply a lightweight post-processing procedure: we crop the reconstruction to the target row region, remove background structures and supports, and filter remaining outliers using statistical outlier removal~\cite{zhou2018open3d}.
Finally, to enable cross-session comparison in a consistent coordinate system, we align all sessions to the first capture day using a landmark-guided ICP procedure, where landmarks are manually selected on static greenhouse structures (e.g., rail supports) to ensure a high-quality benchmark.
These manual steps are used only to construct reliable reference models for simulation-based evaluation.
In the real-world experiments, a common rail-based frame is established through landmark-based odometry initialization, followed by automatic non-rigid registration.

For evaluation across sessions, we construct test cases from the four consecutive session pairs (Sep.~15$\rightarrow$18, 18$\rightarrow$22, 22$\rightarrow$25, and 25$\rightarrow$29), which are 3--4 days apart and reflect a practical greenhouse monitoring cadence.
Unlike the individual-plant protocol, we do not apply additional global rotations or sweep initial viewpoints, since reconstruction starts from a fixed platform configuration.
In total, this yields $2$ (crop types) $\times\,4$ (session pairs) $=8$ test cases.
Since tomato and cherry tomato exhibit similar canopy morphology and our study focuses on structure-level reconstruction rather than cultivar-specific traits, we report results aggregated across the two crop types unless stated otherwise.

\begin{figure}[!t]
\centering
\includegraphics[width=0.9\columnwidth]{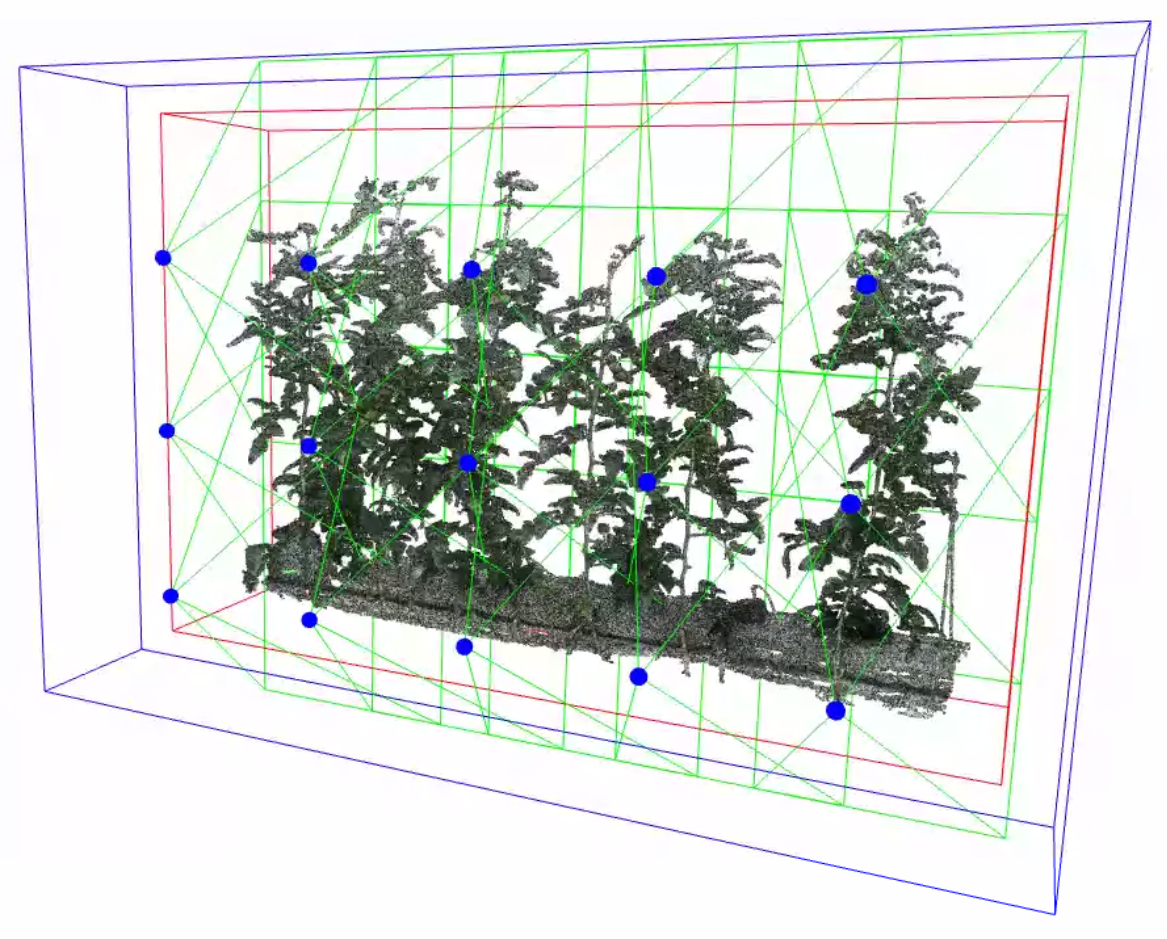}
\caption{
Simulated passive mapping warm start in the greenhouse-row setting based on oriented bounding boxes (OBBs).
The plant OBB (red) bounds the crop structure, and the view OBB (blue) approximates the accessible camera workspace on the aisle/rail side.
We place 15 fixed viewpoints inside the view OBB; their fields of view (green) are arranged to collectively cover the plant OBB, following the sensing geometry (camera distance and vertical spacing) of representative greenhouse monitoring platforms (e.g.,~\cite{smitt2021pathobot}).
For visualization, we overlay the viewpoints on the reference (ground-truth) point cloud.
}
\label{fig_passive_fov}
\vspace{-0.2cm}
\end{figure}

\begin{figure*}[!t]
\centering
\includegraphics[width=1.0\textwidth]{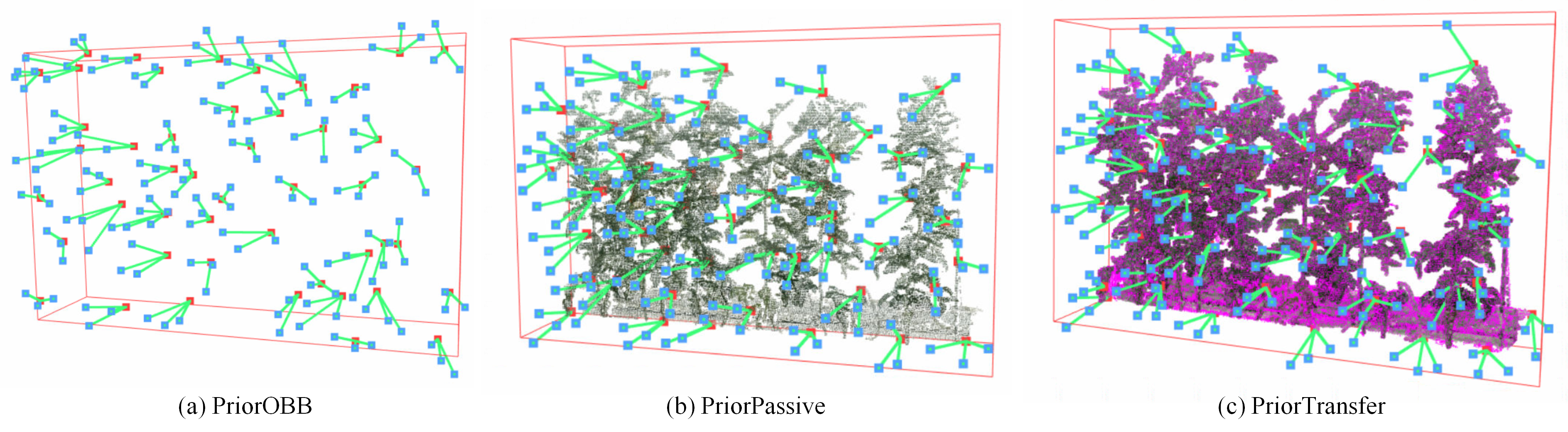}
\caption{
Candidate 6D view generation under the prior ladder for look-at sampling.
(a) PriorOBB: look-at targets are sampled inside the plant OBB (red).
(b) PriorPassive: look-at targets are sampled from surface frontiers of the passive coarse reconstruction (raw plant colors).
(c) PriorTransfer: look-at targets are sampled on the transferred surface obtained from the previous-session reconstruction (purple), excluding regions already observed in the coarse reconstruction.
In all cases, camera positions (blue) are drawn from the same rail-side view OBB and assigned to look-at targets using a $k$-NN rule with balanced allocation to form 6D candidates (green directions).
For clarity, the visualization shows a subset of 50 targets and $k=3$ assignments; experiments use 100 targets with $k=8$ (800 candidates) per strategy.
This figure is meant to visualize the three sampling strategies; quantitative comparisons are reported in Table~\ref{tab_uplimit} and subsequent experiments.
}
\label{fig_prior_ladder}
\end{figure*}

\subsubsection{View Space Configuration and Parameterization}
We define the feasible view space in the greenhouse-row setting based on oriented bounding boxes (OBBs) of the reconstructed row.
Across sessions, the row geometry spans roughly $3.6\,\mathrm{m}$ in length, $0.7$--$0.8\,\mathrm{m}$ in width, and $2.3$--$2.5\,\mathrm{m}$ in height.
We use these OBBs to construct (i) a plant OBB that bounds the crop structure and (ii) a view OBB that approximates the accessible camera workspace on the aisle side (consistent with typical greenhouse monitoring platforms~\cite{smitt2021pathobot}).
As illustrated in Fig.~\ref{fig_passive_fov}, the plant OBB follows the crop-row orientation, while the view OBB defines a neighboring camera-sampling region on the aisle/rail side.
These boxes encode the row geometry and the spatial arrangement of the monitoring platform.

Within the view OBB, we use Poisson-disk sampling to obtain a nearly uniform pool of 1,000 camera positions shared by all compared methods.
Unlike the object-centric individual-plant setting, greenhouse-row reconstruction requires explicit planning of viewing directions under restricted fields of view.
Therefore, we parameterize each view by a 3D camera position together with a 3D look-at target within the plant OBB.
We sample 100 look-at targets and assign eight camera positions from the shared position pool to each target using a $k$-nearest-neighbor rule with balanced round-robin allocation.
This yields 800 candidate 6D views, each defined by a camera position and its associated look-at target.
The sampling budget is fixed across the compared look-at strategies to isolate the effect of geometric guidance.

For oracle construction and evaluation, the reference view space $\mathcal{V}_{\mathrm{ref}}$ combines the 1,000 sampled camera positions with all surface voxels of the reference model as potential look-at targets.
We compute the oracle-visible set $\Omega$ in Eq.~\eqref{eq:Omega} by OctoMap ray casting from the sampled positions toward the surface targets.
This procedure is performed offline (taking on the order of hours per scene in our implementation) and captures the maximum observable surface under the feasible view space, thereby avoiding penalizing methods for inherently unobservable regions (e.g., back-facing structures).
The OctoMap resolution used by the virtual sensor (i.e., the voxel size for coverage evaluation on the reference model) is set to 0.01\,m to match the meter-scale greenhouse setting.

\subsubsection{Setting-Specific Implementation}
In the greenhouse-row setting, we obtain a coarse reconstruction via a passive mapping warm start inspired by representative greenhouse monitoring systems~\cite{smitt2021pathobot}, which use multiple fixed cameras for routine data acquisition.
To simulate such passive sensing in our view-planning simulator, we predefine 15 fixed viewpoints within the view OBB and choose their placements such that their combined fields of view cover the plant OBB.
We fuse the point clouds rendered from these 15 viewpoints and use their union as the warm-start reconstruction for subsequent non-rigid registration.
Fig.~\ref{fig_passive_fov} visualizes the simulated passive mapping setup.

For non-rigid registration, we use the sparse locally rigid warp-field implementation adopted from~\cite{lobefaro2024spatio}, with control-node downsampling resolution $r_{\mathrm{graph}}=0.15$\,m.
Each source point is associated with its four nearest control nodes.
The dense deformation field is obtained by Gaussian-weighted interpolation of node rotations and translations, with the blended rotation projected onto $\mathrm{SO}(3)$.
The data term in Eq.~\eqref{eq:nricp_obj} uses point-to-plane residuals between deformed source points and their nearest-neighbor target correspondences.
Correspondences farther than 0.5\,m are excluded.
At each ICP iteration, the spatial stiffness regularizer applies a component-wise Huber penalty to differences between neighboring control nodes' incremental rotation and translation parameters.
We set $\lambda_{\mathrm{data}}=1$ and $\lambda_{\mathrm{reg}}=2000$.
The Gaussian weighting scales for point-to-node interpolation and graph regularization are $0.5r_{\mathrm{graph}}$ and $3r_{\mathrm{graph}}$, respectively.
The Huber boundary parameter is 0.01.
We perform up to 15 ICP iterations, with up to 10 Gauss--Newton iterations per update.
The resulting sparse linear systems are solved using conjugate gradient with up to 500 iterations.

For inflation, we use a voxel-based implementation to avoid exhaustive grid traversal over the larger scene.
We voxelize the current coarse reconstruction and the aligned previous-session reconstruction on a common 0.01\,m grid and dilate their occupied-voxel sets using spherical neighborhoods with radii $\gamma_{\mathrm{near}}=0.01$\,m and $\gamma_{\mathrm{far}}=0.01$\,m, respectively.
We retain voxels in the current-map dilation that lie outside the history-map dilation, excluding voxels already occupied by the current map.
The retained voxel centers form the added inflation set $\mathcal{I}$.
This inflation grid remains fixed across the different covering resolutions.

For set covering, we voxelize the inflated approximation at covering resolution $r_{\mathrm{cover}}\in\{10,5,4,3\}\,\mathrm{cm}$ when constructing $\mathcal{P}_{\mathit{surf}}$ and the associated visibility sets.

For path planning, we compute a hierarchical global path over the selected views to approximate execution effort under restricted camera workspaces.
Because the manipulator workspace is limited (e.g., a UR5-class arm), we split the scan into upper and lower height bands and solve the global path planning within each band, with a connection view to concatenate the two paths.
The open-TSP subproblem within each height band is solved using Gurobi~\cite{gurobi2021gurobi}.

\subsubsection{Baselines}
We consider two baseline families: (i) a sequential NBV baseline with online look-at sampling, and (ii) one-shot baselines based on a prior ladder that isolates the effect of different levels of geometric guidance on look-at generation.
Here, prior refers to the amount of geometric information used to guide sampling, rather than a learned or probabilistic prior.

\textbf{(i) Block-wise sequential NBV.}
We reimplement the NBV pipeline of Zaenker~\etalcite{zaenker2021viewpoint} with information gain over unknown voxels and sampling online look-at targets from surface frontiers (unknown voxels adjacent to both free and occupied neighbors).
In our greenhouse setup, the manipulator workspace is limited, making long-range sequential NBV prone to inefficient back-and-forth motions.
Inspired by block-wise planning strategies in~\cite{zaenker2023graph,jose2025go}, we therefore adopt a block-wise execution scheme: we partition the view OBB into 10 blocks (5 in an upper height band and 5 in a lower band) and sequentially move the manipulator to each block to perform NBV selection locally.
We refer to this baseline as Block-Wise NBV.

\textbf{(ii) One-shot baselines: prior ladder for look-at generation.}
To analyze how prior strength affects look-at sampling, we design a three-level prior ladder that generates candidate 6D views from the same view OBB but with increasingly informative priors:
\begin{enumerate}
    \item[a)] \textbf{OBB-only (PriorOBB).} We sample look-at targets by Poisson-disk sampling inside the plant OBB.
    \item[b)] \textbf{OBB + passive (PriorPassive).} We extract surface frontiers from the coarse reconstruction obtained by passive mapping and sample look-at targets by farthest-point sampling on this set.
    \item[c)] \textbf{OBB + passive + previous-session transfer (PriorTransfer).} We sample look-at targets via farthest-point sampling on the deformed and inflated prior surface, excluding surface points that are already observed in the coarse reconstruction.
\end{enumerate}
Fig.~\ref{fig_prior_ladder} visualizes the candidate 6D view generation for the three prior levels.

Note that the OBB-only and passive-only strategies do not provide a reliable full-surface geometric prior and therefore cannot directly support our set-cover formulation.
We thus use random selection over the generated candidate views as the corresponding one-shot baselines: PriorOBBRandom, PriorPassiveRandom, and PriorTransferRandom.
Our method uses the previous-session transfer-prior strategy (c) together with set-cover-based view selection.

\begin{table}[!t]
\centering
\resizebox{1.0\columnwidth}{!}{%
\begin{tabular}{|c|c|c|c|}
\hline
Look-At Sampling Strategy       & PriorOBB     & PriorPassive & PriorTransfer \\
\hline
Max Reach of $\Omega$ (\%) & 94.48 ± 1.17 & 97.21 ± 0.90  & \textbf{97.34} ± 0.77 \\
\hline
\end{tabular}
}
\caption{Effect of geometric guidance on reachable coverage.
We report the maximum reachable fraction of the oracle visible set $\Omega$ under three look-at sampling strategies (OBB-only, passive, and previous-session transfer), using the same 1000 candidate positions (mean $\pm$ std across test cases).
Compared to OBB-only, passive and transfer-based guidance increases the reachable coverage upper bound, indicating improved look-at sampling rather than a lack of candidate-position density.
Moreover, passive and transfer yield similar upper bounds, suggesting that the primary role of previous-session transfer in this setting is not to expand the theoretically observable surface, but to provide a more reliable planning geometry for consistent visibility evaluation and set-cover selection.
}
\label{tab_uplimit}
\end{table}

\begin{figure*}[!t]
\centering
\includegraphics[width=1.0\textwidth]{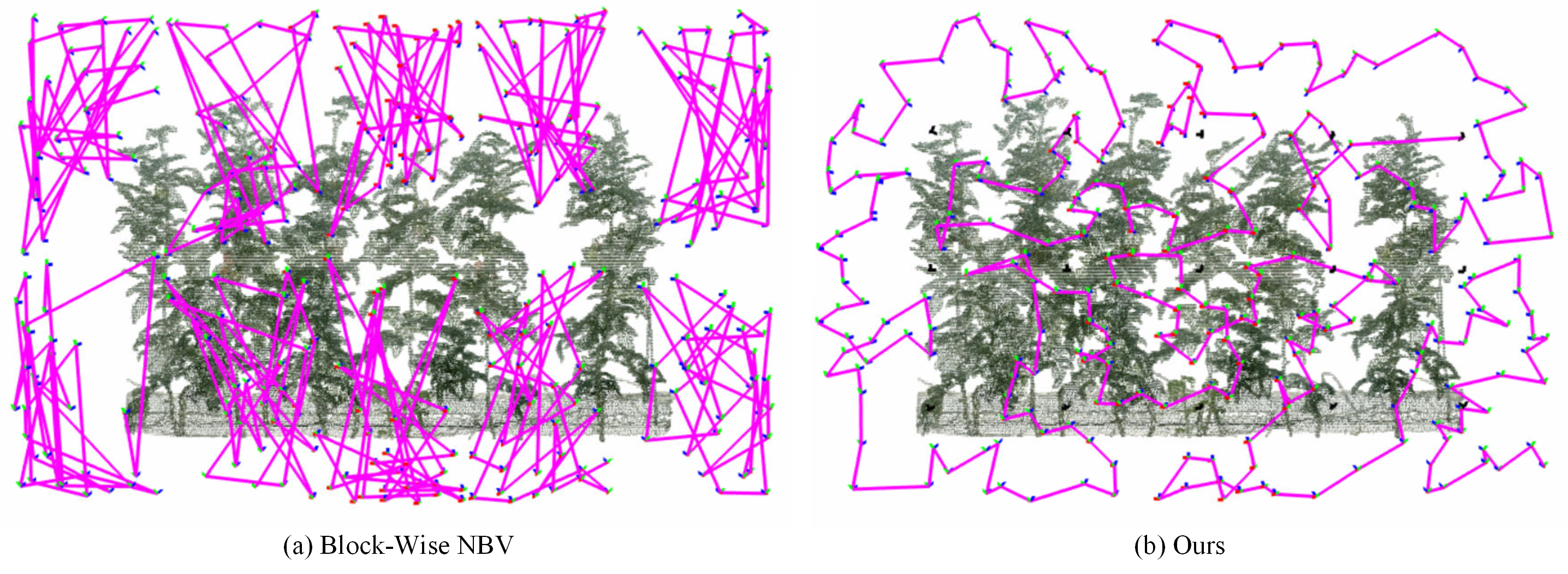}
\caption{
Qualitative comparison of execution paths in the greenhouse-row setting.
Block-wise NBV (a) and Ours (b) are shown at a matched coverage level: NBV reaches 92.42\% surface coverage with 300 views, while Ours (5\,cm) reaches 92.68\% with 280 views.
Block-wise NBV executes views sequentially in its native order within each block, resulting in frequent back-and-forth motions and a long travel path.
Despite similar coverage, our method yields slightly fewer view counts and a substantially more compact path due to the global path planning.
}
\label{fig_compare_greenhouse}
\end{figure*}

\begin{table}[!t]
\centering
\resizebox{1.0\columnwidth}{!}{%
\begin{tabular}{|c|c|c|c|}
\hline
Method & \makecell[c]{Number\\of Views} & \makecell[c]{Surface\\Coverage (\%)} & \makecell[c]{Path\\Length (m)}     \\
\Xhline{1.2pt}
\makecell[c]{Coarse Reconstruction}  & 15             & 44.48 ± 1.09 & \textbf{2.58} ± 0.02  \\
\Xhline{1.2pt}
\multirow{3}{*}{Block-Wise NBV}                 & 100            & 73.73 ± 2.46 & 64.94 ± 4.14  \\
                                                & 300            & 90.06 ± 2.12 & 174.20 ± 6.95 \\
                                                & 500            & 94.70 ± 1.60 & 279.02 ± 7.68 \\

\Xhline{1.2pt}
\multirow{3}{*}{PriorOBBRandom}                 & 100            & 61.44 ± 3.25 & 28.78 ± 0.85 \\
                                                & 300            & 82.94 ± 2.11 & 60.25 ± 1.28 \\
                                                & 500            & 89.69 ± 1.44 & 87.09 ± 1.22 \\
\Xhline{1.2pt}
\multirow{3}{*}{PriorPassiveRandom}             & 100            & 71.99 ± 2.69 & 28.93 ± 0.74 \\
                                                & 300            & 88.15 ± 1.87 & 57.99 ± 0.77 \\
                                                & 500            & 93.45 ± 1.17 & 83.04 ± 0.67 \\
\Xhline{1.2pt}
\multirow{3}{*}{PriorTransferRandom}            & 100            & 72.43 ± 2.20 & 28.91 ± 0.86 \\
                                                & 300            & 88.37 ± 1.40 & 57.98 ± 0.59 \\
                                                & 500            & 93.57 ± 1.07 & 83.07 ± 0.48 \\
\Xhline{1.2pt}
\multirow{4}{*}{Ours}                    & 71.3 ± 7.5 (10\,cm)   & 69.89 ± 2.58 & 23.73 ± 1.64 \\
                                                & 255.9 ± 19.1 (5\,cm)  & 90.55 ± 1.53 & 53.21 ± 2.28 \\
                                                & 345.8 ± 18.9 (4\,cm)  & 93.26 ± 1.30 & 64.54 ± 2.36 \\
                                                & 476.0 ± 17.1 (3\,cm)  & \textbf{95.46} ± 1.01 & 80.28 ± 2.19 \\
\Xhline{1.2pt}
\end{tabular}
}
\caption{Coverage--effort trade-off in the greenhouse-row setting.
We report oracle-visible surface coverage, executed views, and path length for a passive warm start (Coarse Reconstruction), a sequential baseline (Block-Wise NBV), one-shot baselines using different look-at sampling strategies with random selection (PriorOBB/Passive/Transfer), and our method.
Block-Wise NBV is sequential, so results are shown at fixed view budgets (100/300/500), whereas our method solves a one-shot set-cover problem and outputs a variable number of views.
All one-shot methods, including ours, use the same global path planner, while look-at sampling and view selection follow their respective strategies.
The four Ours rows correspond to different covering resolutions (voxel sizes 10/5/4/3\,cm), yielding multiple operating points on the coverage--effort frontier.
For methods that use random selection, the experiments were repeated five times for evaluation.
For methods that use the passive warm start, the reported views and path lengths include the 15 fixed passive viewpoints.
Values are mean $\pm$ std across test cases.
Overall, our method achieves substantially higher coverage than random one-shot baselines at comparable path lengths, and it reaches NBV-level coverage with dramatically shorter path lengths.
}
\label{tab_greenhouse}
\end{table}

\subsubsection{Analysis on Upper-Limit Surface Coverage}
Table~\ref{tab_uplimit} reports an upper-bound analysis of reachable visible surface coverage under the three look-at sampling strategies in our prior ladder.
Using the same set of 1,000 candidate positions, the OBB-only strategy achieves a noticeably lower maximum reach (94.48\%), indicating that coarse spatial priors alone are insufficient to consistently orient the sensor toward occluded structures in greenhouse rows.
Importantly, OBB-only still reaches 94.48\% despite using a dense candidate set, suggesting that the remaining gap is mainly attributable to limited look-at guidance rather than insufficient candidate density.
In contrast, incorporating the passive warm-start substantially increases the upper bound to 97.21\%, showing that surface-frontier cues extracted from passive mapping already provide strong guidance for look-at selection and greatly reduce the fraction of hard-to-see regions.
Notably, the transfer strategy yields a similar upper bound (97.34\%) to the passive strategy, suggesting that its main role in this setting is not to expand the theoretically observable surface, but to provide a more reliable planning geometry for consistent visibility evaluation and set-cover selection.
The high reachable fractions obtained with passive and transfer guidance support the adopted candidate-generation budget for this benchmark, retaining access to over 97\% of the oracle-visible surface defined over the shared position pool.

\subsubsection{Evaluation of View Planning Performance}
We evaluate the coverage--effort trade-off in the greenhouse-row setting by comparing a sequential baseline (Block-Wise NBV), one-shot baselines under the prior ladder, and our method.
Table~\ref{tab_greenhouse} summarizes oracle-visible surface coverage and acquisition cost (executed views and path length).

\textbf{Sequential baseline (Block-Wise NBV).}
Block-wise NBV achieves steadily increasing coverage as the view budget grows (e.g., 90.06\% at 300 views and 94.70\% at 500 views), reflecting the benefit of online refinement.
However, this gain comes with a dramatic increase in execution cost: even with block-wise execution, the traveled path grows from 64.94\,m (100 views) to 174.20\,m (300 views) and further to 279.02\,m (500 views), far exceeding all one-shot methods.
This indicates that locally greedy sequential decisions remain poorly matched to execution-constrained greenhouse-row monitoring, where redundant back-and-forth motions translate into large travel overhead.

\textbf{One-shot baselines under the prior ladder.}
Random one-shot selection already yields substantially shorter paths than NBV (about 29--58\,m for 100--300 views), highlighting the advantage of selecting views in a batch and routing them globally.
Within this family, stronger look-at priors consistently improve coverage at a fixed view budget: PriorOBBRandom lags behind, whereas Passive- and Transfer-guided look-at sampling achieves markedly higher coverage (e.g., 88.15\% and 88.37\% at 300 views).
These results support our hypothesis that, in greenhouse rows, the main performance gains come from improved look-at guidance toward under-observed structures, rather than from position sampling alone.

\textbf{Ours: coverage--cost operating points.}
Our method further pushes the coverage--cost frontier by replacing random selection with set-cover-based view selection.
By adjusting the voxel size in the covering stage, our method yields multiple operating points that trade additional coverage for additional resources.
Fig.~\ref{fig_compare_greenhouse} provides a qualitative comparison at a matched coverage level, where our method achieves similar coverage with fewer views and a substantially more compact path than Block-Wise NBV.
At a balanced resolution (4\,cm), our method reaches near-high coverage (93.26\%) with a 64.54\,m path using 345.8 views, while Block-Wise NBV attains a comparable coverage level only with substantially higher execution cost (e.g., 500 views and 279.02\,m of travel for 94.70\% coverage).
At a finer resolution (3\,cm), our method achieves the highest coverage in the table (95.46\%) while still maintaining a much shorter path than the sequential baseline.
Overall, these results show that one-shot planning with principled look-at assignment and set-cover selection can achieve high reconstruction completeness with substantially lower acquisition cost, and that the covering resolution provides a practical knob to control the coverage--cost trade-off.

\begin{table}[!t]
\centering
\resizebox{0.8\columnwidth}{!}{%
\begin{tabular}{|c|c|c|}
\hline
Previous-Session Input & \makecell[c]{Number\\of Views} & \makecell[c]{Surface\\Coverage (\%)} \\
\Xhline{1.2pt}
Previous GT    & $345.5 \pm 21.8$ & $93.01 \pm 1.29$ \\
\hline
Previous Dense & $346.8 \pm 19.2$ & $93.07 \pm 1.28$ \\
\hline
Previous Ours  & $349.7 \pm 19.7$ & $93.19 \pm 1.22$ \\
\hline
\multicolumn{3}{|c|}{Surface coverage: exact Friedman $p=0.366$} \\
\Xhline{1.2pt}
\end{tabular}
}
\caption{
Effect of the previous-session reconstruction source in greenhouse-row simulation at a covering resolution of 4\,cm.
Previous GT uses the previous session's ground-truth geometry, Previous Dense uses its oracle-visible point cloud constructed from all 1,000 Poisson-disk-sampled camera positions and all voxelized GT surface targets, and Previous Ours uses its planned reconstruction.
All conditions share the same current-session passive observations, planning settings, and oracle-visible surface coverage reference.
Reported view counts include the 15 fixed passive viewpoints.
Values are mean $\pm$ sample standard deviation across six paired cases from two greenhouse rows and three session pairs.
An exploratory exact Friedman test detects no significant surface-coverage difference among the three inputs ($p=0.366$, $\alpha=0.05$).
The reported averages differ slightly from those in Table~\ref{tab_greenhouse}, as this comparison is restricted to six paired cases rather than the full greenhouse-row evaluation set.
Overall, using the previous session's planned reconstruction achieves comparable surface coverage to Previous GT and Previous Dense with only a modest increase in the number of acquired views, supporting its use in repeated monitoring.
}
\label{tab:previous_session_source}
\end{table}

\begin{table*}[!t]
\centering
\resizebox{1.0\textwidth}{!}{%
\begin{tabular}{|c|c|c|c|c|c|c|c|c|c|}
\hline
\multirow{2}{*}{$\sigma_t$ (m)} & \multirow{2}{*}{$\sigma_r$ (deg)} & \multirow{2}{*}{ICP} & \multirow{2}{*}{Chamfer Distance (mm)} & \multicolumn{3}{c|}{F-score ($\tau=0.01 \,m$)}               & \multicolumn{3}{c|}{F-score ($\tau=0.02\,m$)}               \\
\cline{5-10}
                          &                           &                      &                          & Precision    & Recall       & $F_1$           & Precision    & Recall       & $F_1$           \\
\Xhline{1.2pt}
0                         & 0                         & -                   & \textbf{5.44 }± 0.13              & \textbf{100.0} ± 0.0    & 87.3 ± 1.4 & \textbf{93.2} ± 0.8 & \textbf{100.0} ± 0.0    & 97.9 ± 0.6 & \textbf{98.9} ± 0.3 \\
\Xhline{1.2pt}
0.002                     & 0.5                       & OFF                   & 7.69 ± 0.15              & 57.8 ± 2.0   & 91.8 ± 1.3 & 70.9 ± 1.4 & 93.7 ± 0.9 & 98.6 ± 0.5 & 96.1 ± 0.5 \\
0.005                     & 1                         & OFF                   & 9.89 ± 0.27              & 40.0 ± 1.9   & \textbf{93.1} ± 1.6 & 56.0 ± 1.7   & 76.7 ± 2.3 & 99.0 ± 0.5   & 86.4 ± 1.4 \\
0.01                      & 2                         & OFF                   & 13.87 ± 0.47             & 27.3 ± 1.6 & 92.2 ± 1.5 & 42.1 ± 1.8 & 55.3 ± 2.4 & \textbf{99.3} ± 0.3 & 71.0 ± 2.0     \\
\Xhline{1.2pt}
0.002                     & 0.5                       & ON                  & 5.89 ± 0.13              & 99.2 ± 0.5 & 85.7 ± 1.6 & 91.9 ± 0.8 & 100.0 ± 0.0    & 97.7 ± 0.6 & 98.8 ± 0.3 \\
0.005                     & 1                         & ON                  & 6.03 ± 0.18              & 98.9 ± 0.4 & 85.0 ± 1.8   & 91.4 ± 1.1 & 100.0 ± 0.0    & 97.6 ± 0.6 & 98.8 ± 0.3 \\
0.01                      & 2                         & ON                  & 6.08 ± 0.16              & 98.7 ± 1.1 & 85.0 ± 1.8   & 91.4 ± 1.0   & 99.7 ± 0.6 & 97.6 ± 0.6 & 98.7 ± 0.2 \\
\Xhline{1.2pt}
\end{tabular}
}
\caption{Pre-deployment pose-noise robustness on the greenhouse-row setting of our method (4\,cm).
We perturb active-view camera poses with translational noise $\sigma_t$ (m) and rotational noise $\sigma_r$ (deg), and report geometric fidelity against the ground-truth surface using symmetric Chamfer Distance and F-scores at $\tau\in\{0.01,0.02\}$\,m (precision, recall, and $F_1$).
Results are shown with and without lightweight ICP refinement (ICP column) and are reported as mean $\pm$ std over all test cases.
Pose noise without refinement primarily degrades precision and increases Chamfer distance, whereas lightweight ICP largely recovers the geometry metrics to near the no-noise baseline across the tested noise levels.
}
\label{tab_noise}
\end{table*}

\begin{figure*}[!t]
\centering
\includegraphics[width=0.75\textwidth]{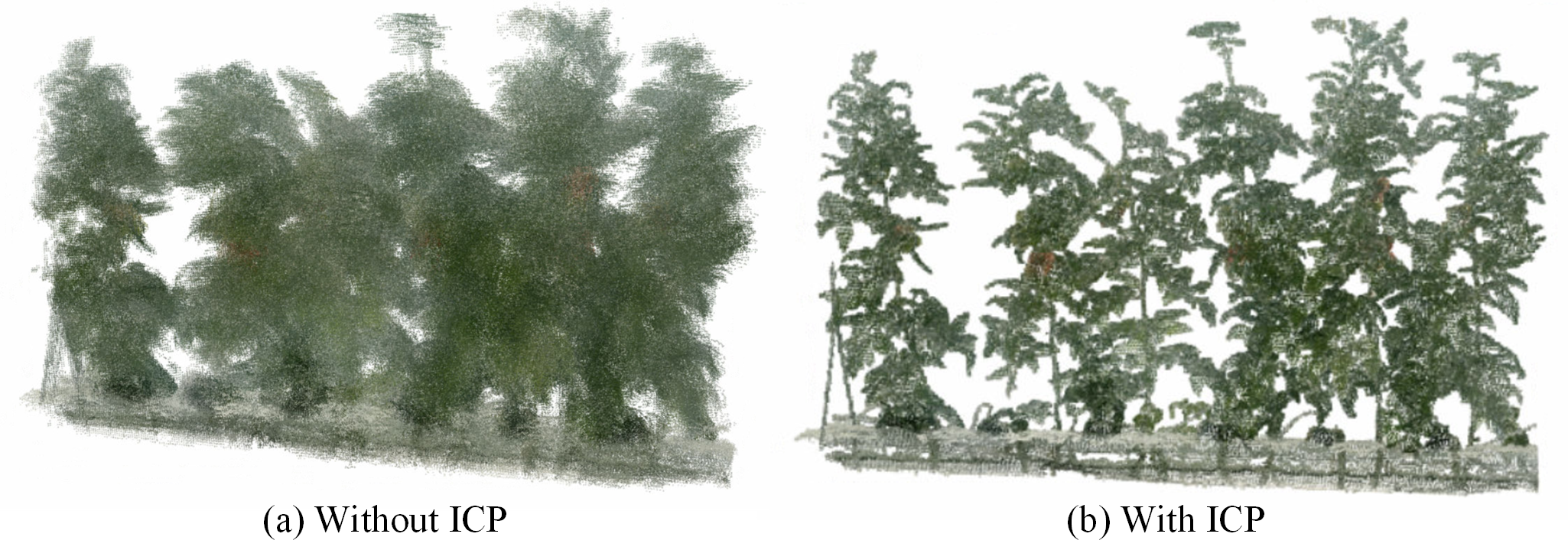}
\caption{
Qualitative pose-noise robustness example in the greenhouse-row setting (our method at 4\,cm) shown at $(\sigma_t,\sigma_r)=(0.01\,\mathrm{m},2^\circ)$.
We perturb the executed active-view poses and fuse the resulting observations.
Without refinement (a), pose noise leads to drift and severe blurring/ghosting in the fused point cloud.
With lightweight ICP refinement during fusion (b), the reconstruction is largely re-aligned, recovering sharp plant structures and improving geometric fidelity.
}
\label{fig_pose_noise}
\end{figure*}

\textbf{Planning-Time Analysis.}
We report the computation time of different planners for generating view sets and their execution orders in the greenhouse-row setting.
All timings in this comparison use the same multithreaded CPU platform, without CUDA acceleration.
For the sequential baseline (Block-Wise NBV), planner computation takes about 3.5\,s per iteration, amounting to roughly 1750\,s for 500 iterations/views.
For our method, non-rigid deformation takes about 3--8\,s, set-cover optimization about 1\,s, and global path planning about 2--10\,s.
The dominant cost is OctoMap ray casting for visibility evaluation before set covering, which takes approximately 250/700/1200/2000\,s for covering voxel sizes of 10/5/4/3\,cm, respectively.
Since our method generates the view plan once per monitoring session, this computation is performed before active acquisition without requiring replanning at every observation.
For a consistent CPU-based comparison, we report unaccelerated timings here and evaluate CUDA-accelerated visibility computation separately in the real-world deployment experiments (Table~\ref{tab:cuda_visibility}).

\subsubsection{Summary}
Overall, the greenhouse-row experiments highlight two key takeaways.
First, under execution-constrained crop-row monitoring, globally planned one-shot view selection with global path planning achieves a substantially better coverage--effort trade-off than sequential NBV baselines by reducing redundant back-and-forth motions.
Second, at the meter scale with restricted fields of view, transferring the previous-session reconstruction improves reconstruction mainly by guiding 3D look-at decisions toward previously under-observed structures, rather than by position sampling alone.

\subsection{Effect of Previous-Session Reconstruction Source}

The preceding simulation evaluation uses a relatively comprehensive, manually acquired point cloud from the previous session (denoted as GT) as the reference input to our reconstruction-transfer pipeline.
Although this point cloud does not capture all plant surfaces, it provides a consistent reference for evaluating the view-planning method.
In repeated robotic monitoring, however, the previous-session input is reconstructed from observations acquired during the previous session.
We therefore examine how replacing the manually acquired reference with reconstructions obtained through dense sampling or our planned acquisition affects subsequent view planning.

\subsubsection{Protocol}
We compare three sources of previous-session geometry:
\begin{enumerate}
    \item[a)] \textbf{Previous GT.} We use the ground-truth point cloud of the previous session as the reference input, consistent with the preceding simulation evaluation.
    \item[b)] \textbf{Previous Dense.} We use the previous session's oracle-visible point cloud as the dense reconstruction.
    This point cloud is constructed as the union of GT surface voxels visible from all 1,000 Poisson-disk-sampled camera positions in the shared view space, using ray casting toward every surface target of the voxelized GT model.
    This follows the same oracle construction used to define the surface-coverage denominator, applied to the previous session.
    \item[c)] \textbf{Previous Ours.} We use the reconstruction obtained from the previous session's passive observations and active views selected by our method.
\end{enumerate}
The first monitoring session (Sep.~15) provides the initial reference geometry, so a reconstruction produced by our planned acquisition is first available for Sep.~18.
The comparison therefore covers three subsequent session pairs: Sep.~18$\rightarrow$22, Sep.~22$\rightarrow$25, and Sep.~25$\rightarrow$29, 2025.
Across the two greenhouse rows, this yields six paired cases.
For each current scene, only the previous-session input is replaced; the current-session passive observations, planning settings, and coverage-evaluation reference remain unchanged.
Each current-session reconstruction is built from newly acquired observations rather than by accumulating historical geometry, so this comparison assesses one-step prior substitution across session pairs.
Results are reported as mean and sample standard deviation across these six cases.
We additionally perform an exploratory exact Friedman test on paired current-session surface coverage, enumerating all within-case permutations of the three source labels at a significance level of $\alpha=0.05$.

\subsubsection{Evaluation of Previous-Session Reconstruction Source}
We evaluate our method at a covering resolution of 4\,cm, which provides a balanced trade-off between coverage and acquisition cost in the greenhouse-row experiments.
As shown in Table~\ref{tab:previous_session_source}, Previous GT, Previous Dense, and Previous Ours achieve mean oracle-visible surface coverage of 93.01\%, 93.07\%, and 93.19\%, respectively, with corresponding mean view counts of 345.5, 346.8, and 349.7.
Relative to Previous GT, Previous Ours requires 4.2 additional views on average, while the largest paired coverage decrease is only 0.06 percentage points.
The exact Friedman test detects no statistically significant difference in current-session surface coverage among the three inputs across the tested cases ($p=0.366$).
Because the six cases contain repeated observations of only two greenhouse rows, this test is exploratory and does not establish statistical equivalence.

The slightly higher mean coverage with Previous Dense and Previous Ours may reflect their focus on previously observable surfaces.
By excluding reference surfaces that were not observed in the previous session, these inputs may bias transferred look-at sampling toward regions more likely to be observable from the feasible view space.
These results indicate comparable coverage with a modest change in acquisition effort across the tested cases, supporting the use of a previous planned reconstruction as the input for subsequent monitoring.

\subsection{Pre-Deployment Pose-Noise Robustness}

In practical deployment, our greenhouse data acquisition is performed with an eye-in-hand RGB-D sensor mounted on a manipulator, where each planned 6D view is executed by the arm while the rail-based base provides coarse repositioning along the crop row.
In this setup, pose inaccuracies associated with hand-eye calibration, manipulator execution, and platform motion can introduce misalignment during multi-view fusion.
We therefore conduct a pre-deployment pose-noise robustness study to quantify how execution errors affect reconstruction quality and to evaluate whether lightweight ICP refinement can recover geometric fidelity.

\subsubsection{Protocol}
To assess reconstruction sensitivity to pose inaccuracies, we perturb the transformations used to place active-view point clouds in the common coordinate frame with zero-mean Gaussian noise.
The simulated observations are kept fixed to isolate misalignment during fusion.
Specifically, we add translational noise with standard deviation $\sigma_t$ (in meters) to the camera position and rotational noise with standard deviation $\sigma_r$ (in degrees) to the camera orientation.
We test three noise levels, $(\sigma_t,\sigma_r)\in\{(0.002,0.5),(0.005,1.0),(0.01,2.0)\}$, as well as the noise-free case.
Passive warm-start observations are kept noise-free to isolate the impact of execution errors during active acquisition.
For each noise level, we compare reconstruction with and without lightweight ICP refinement during fusion.
With ICP enabled, each perturbed active-view cloud is registered to the accumulated current-session map, initialized from the passive observations, before being fused.
Registration uses 0.02\,m voxel downsampling, a maximum correspondence distance of 0.05\,m, and up to 50 iterations.
The estimated transformation is accepted only if ICP converges and its mean-squared fitness score is below $(0.02\,\mathrm{m})^2$; otherwise, the unrefined cloud is fused.

\subsubsection{Evaluation of Pose-Noise Robustness}
We evaluate pose-noise robustness at a representative operating point of our method by fixing the covering resolution to 4\,cm, which provides a balanced trade-off between coverage and acquisition cost in the greenhouse-row experiments.
We then perturb the executed active-view poses according to the protocol above and report geometric fidelity in Table~\ref{tab_noise}.

\textbf{Effect of pose noise without refinement.}
Without ICP, increasing pose noise leads to a clear degradation in geometric accuracy: Chamfer Distance increases monotonically and the F-score at $\tau=0.01$\,m drops sharply.
This behavior is consistent with the fact that small pose perturbations primarily introduce misalignment during fusion, causing reconstructed points to deviate beyond a tight distance threshold.

\textbf{Lightweight ICP largely recovers geometric fidelity.}
With ICP enabled, reconstruction quality becomes largely insensitive to the tested noise levels.
Fig.~\ref{fig_pose_noise} provides a qualitative example: pose perturbations cause substantial drift and ghosting without refinement, whereas lightweight ICP largely restores alignment and recovers clear plant structures, consistent with the improvements in Chamfer Distance and F-scores in Table~\ref{tab_noise}.
Across all noise settings, ICP reduces Chamfer Distance back toward the noise-free regime and restores high F-scores at both thresholds.
The recovery in F-score is driven primarily by increased precision, while recall is lower after refinement.
Recall remains limited in part by regions of the ground-truth surface that are inherently unobservable from the feasible view space, such as back-facing and inner surfaces.

\textbf{Why ICP works well in our pipeline.}
Two aspects of our pipeline facilitate robust refinement.
First, the passive warm-start map provides a stable initialization for alignment, so the subsequent ICP refinement only needs to correct residual drift introduced by noisy active poses rather than resolve large global misregistrations.
Second, this rigid ICP refinement operates during current-session reconstruction and leaves the precomputed view plan unchanged.
Overall, these results suggest that our method remains practically deployable under realistic execution errors when paired with lightweight ICP refinement.

\section{Real-World Greenhouse Experiments}
\label{sec:real_world}

We organize the real-world experiments to assess the practical deployment of our method for repeated tomato monitoring.
We first address how repeated monitoring can be initialized from a dense bootstrap and continued using reconstructions acquired by the robot in preceding sessions.
We then investigate how robot-specific reachability and motion constraints can be incorporated to obtain executable view plans while retaining the original set-cover formulation.
Finally, we assess the resulting reconstruction gains and computational and motion costs, including the effects of CUDA-accelerated visibility evaluation and joint-space route ordering.

\subsection{Experimental Setup and Protocol}

In this section, we describe the robotic platform and the protocol used for repeated monitoring in the greenhouse.
We also specify the view-space configuration, the previous-session inputs for the three monitoring cases, and the evaluation metrics.

\begin{figure}[!t]
\centering
\includegraphics[width=0.95\columnwidth]{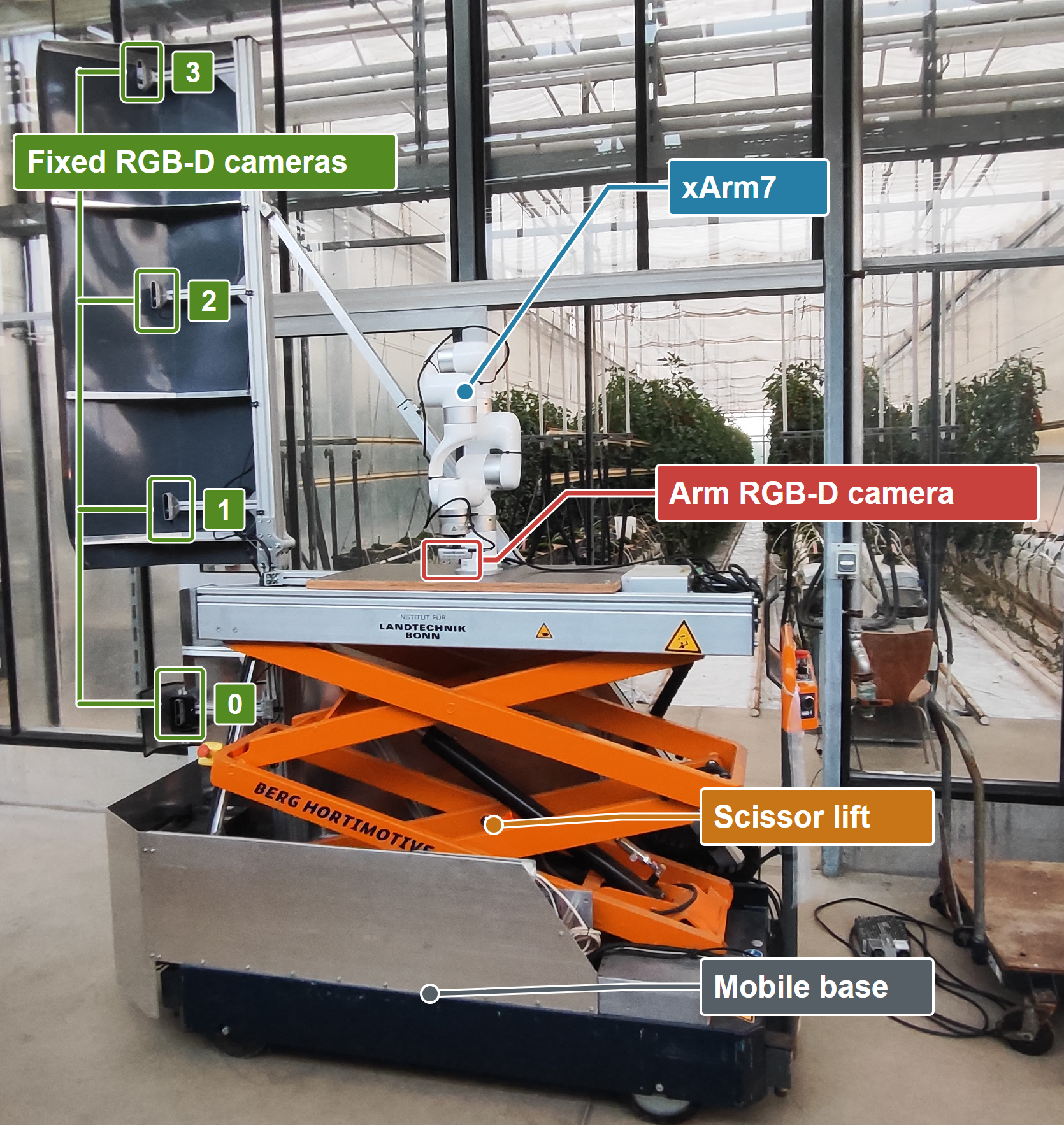}
\caption{
Robotic platform used for the greenhouse experiments.
The platform comprises a mobile base, a scissor lift, an xArm7 manipulator, a fixed RGB-D camera array, and an arm-mounted RGB-D camera.
The lower three fixed cameras (camera0--camera2) are used for passive mapping, following the tomato sensing configuration of PATHoBot~\cite{smitt2021pathobot}.
The arm-mounted camera acquires observations at the planned active viewpoints.
}
\label{fig:real_world_platform}
\end{figure}

\subsubsection{Robotic Platform}
We conduct the experiments using the PATHoBot rail-guided mobile platform~\cite{smitt2021pathobot}, equipped with a scissor lift, as shown in Fig.~\ref{fig:real_world_platform}.
A seven-degree-of-freedom xArm7 manipulator is mounted on the lift platform and carries an Intel RealSense D435 at its end effector.
The manipulator controls the camera position and orientation to acquire observations at the planned active viewpoints.
The platform also carries a fixed array of four RGB-D cameras, numbered camera0--camera3 from bottom to top.
We use the lower three cameras, camera0--camera2, for passive mapping, following the tomato sensing configuration described in~\cite{smitt2021pathobot}.
Rail translation and lift adjustment extend the observation workspace along the tomato row and across different heights, while the manipulator provides local camera positioning within each rail-and-lift configuration.

The arm-mounted camera records RGB-D observations at $1280 \times 720$ resolution, with depth aligned to the color image.
To mitigate unreliable depth measurements under greenhouse illumination, both passive mapping and active acquisition use the RealSense High Accuracy depth preset.
Calibrated fixed-camera extrinsics and the end-effector-to-camera transformation are combined with the recorded platform and arm poses to express observations in a common rail-based coordinate frame.
Mapping and planning run on a PC with an Intel Core i7-14700HX CPU (20 cores), 32\,GB RAM, and an NVIDIA GeForce RTX 4060 Laptop GPU with 8\,GB memory.

\subsubsection{Repeated-Monitoring Protocol}
Before each monitoring session, the robot moves to a fixed rail landmark and resets its rail odometry, establishing a common coordinate reference across sessions.
This initialization is part of the standard startup procedure, after which the odometry remains continuous throughout passive mapping and active acquisition.
The robot first performs a passive rail sweep over approximately 3.5\,m using the lower three fixed RGB-D cameras to obtain a coarse current-session reconstruction.
Passive mapping builds on the implementation of Lobefaro et al.~\cite{lobefaro2024spatio}, which we adapt to use synchronized rail odometry as the global pose reference, with bounded local ICP corrections.
The coarse reconstruction provides the current-session reference for transferring the previous reconstruction.

We then apply the previous-session reconstruction transfer pipeline, including non-rigid registration and conservative inflation, to construct the planning geometry.
Non-rigid registration uses the same sparse locally rigid warp-field implementation and parameter settings as the greenhouse-row simulation.
Inflation uses the same voxel-dilation implementation, 0.01\,m grid spacing, and distance thresholds as the greenhouse-row simulation.
View selection and route planning are performed within the current monitoring session before active acquisition.
The robot executes the precomputed view sequence without replanning during acquisition.
At each selected viewpoint, the robot stops and settles before acquiring a fresh synchronized RGB-D observation, together with the corresponding arm and platform poses.

For reconstruction, active observations acquired at each fixed rail-and-lift configuration are first fused into a block-level point cloud.
Each block cloud is refined by local point-to-point ICP against the current passive map.
The registered block clouds are then fused with the passive observations, followed by 10\,mm voxelization, cropping to the fixed Plant OBB, and statistical outlier removal (SOR) using 30 neighbors and a standard-deviation ratio of 2.0.
The final current-session reconstruction is built exclusively from newly acquired passive and active observations; previous-session geometry is only used to guide planning.
This reconstruction can subsequently serve as the previous-session input for the next monitoring session.

\subsubsection{View Space and Candidate Sampling}
We follow the 6D look-at view parameterization and candidate sampling strategy used in the greenhouse-row evaluation.
Two fixed oriented bounding boxes (OBBs) define the monitored plant region and the camera sampling workspace, respectively.
The Plant OBB spans $3.08 \times 0.45 \times 1.80$\,m and contains six tomato plants, while the camera workspace OBB spans approximately $3.08 \times 0.20 \times 1.66$\,m on the robot-facing side of the row.
Both OBBs are defined once from the greenhouse rail layout, the monitored plant segment, and the robot configuration, and remain unchanged across the three monitoring cases.
Their coordinates are expressed in the common rail-based frame established during initialization.
The current passive map and the previous-session reconstruction are cropped to the Plant OBB for reconstruction transfer and target generation.
The final current-session reconstruction and occupancy evaluation are also restricted to the same Plant OBB.
Although the experiments consider a single row segment, longer rows can be monitored by applying the same pipeline to successive segments without changing the underlying planning algorithm.

The sampling parameters are identical to those used in the greenhouse-row simulation.
We sample 1,000 camera positions using Poisson-disk sampling within the camera workspace OBB.
We select 100 look-at targets from transferred geometry not represented in the current passive map and from the conservative inflation shell.
We assign eight camera positions to each target, yielding 800 raw candidate views.
Each camera position and its associated look-at target define a candidate camera pose.

To distribute acquisition across the monitored region, we use four rail stations and two lift states, corresponding to lift displacements of 0 and 0.8\,m, which define eight acquisition blocks.
The station spacing and lift states are chosen according to the manipulator's reachable workspace, distributing arm motion along the row and across different plant heights.
Candidate sampling uses the full camera workspace OBB, while assignment to rail-and-lift configurations and reachability filtering are performed subsequently.
The reachability-aware procedure described below accounts for joint limits, collision constraints, and cable restrictions before view selection.

\subsubsection{Monitoring Cases and Previous-Session Inputs}
We conduct three repeated-monitoring cases on the same six-plant tomato row segment, using current-session observations acquired on September 30, October 2, and October 5, 2026.
The cases cover initialization from a densely acquired reconstruction, subsequent reuse of reconstructions obtained by our method, and monitoring following routine greenhouse maintenance:
\begin{enumerate}
    \item[a)] \textbf{Case 1: Dense bootstrap.}
    For the September 30 session, we use a previous-session reconstruction acquired on September 29.
    For the September 29 bootstrap acquisition, no earlier reconstruction is available, so look-at targets are sampled from that session's passive map.
    This bootstrap reconstruction combines a passive sweep with a dense acquisition of 569 active views (retained from 800 candidates after reachability filtering and trajectory certification) across the eight rail-and-lift blocks.
    \item[b)] \textbf{Case 2: Reuse of the previous planned reconstruction.}
    For the October 2 session, we use the reconstruction produced by our method in Case 1.
    This previous-session input is built from the September 30 passive observations and the active views selected and acquired by our method.
    \item[c)] \textbf{Case 3: Monitoring after routine harvesting.}
    For the October 5 session, we use the reconstruction produced by our method in Case 2.
    This Monday session follows routine harvesting and therefore includes inter-session plant changes associated with normal greenhouse operations.
    It examines the reuse of a previous planned reconstruction under these practical monitoring conditions.
\end{enumerate}

All previous-session inputs in the real-world experiments are reconstructed from robot-acquired RGB-D observations, without access to GT geometry.
The dense bootstrap in Case 1 provides a practical dense prior and differs from both Previous GT and the oracle-derived Previous Dense used in the simulation-based evaluation.
Cases 2 and 3 follow the same Previous Ours definition as before, using the preceding session's reconstruction formed from passive observations and active views selected by our method.

\subsubsection{Evaluation Metrics}
A complete GT surface model is unavailable for the real-world experiments.
We therefore use relative occupancy gain (OCC gain) to quantify the additional reconstructed geometry obtained through active acquisition.
Let $N_{\mathrm{passive}}$ and $N_{\mathrm{passive+active}}$ denote the numbers of occupied voxels in the passive reconstruction and the final reconstruction, respectively.
We define
\begin{equation}
\mathrm{OCC\ gain}
=
\frac{N_{\mathrm{passive+active}}-N_{\mathrm{passive}}}
{N_{\mathrm{passive}}}
\times 100\%.
\end{equation}
Both reconstructions are evaluated using the same rail-frame-aligned 10\,mm voxel grid, fixed Plant OBB, and outlier-filtering procedure.
This metric measures the increase in observed geometry relative to the passive warm start and quantifies additional occupied-voxel evidence rather than ground-truth surface coverage.

For motion cost, we report planned arm joint travel in radians in place of the camera-space path length used in the simulation-based evaluation.
Camera-space path length provides a platform-independent measure of viewpoint travel, applicable to different sensing platforms, including aerial robots.
For a deployed manipulator, however, a short camera-space displacement can require substantial joint motion, making joint travel more representative of arm actuation effort.
We therefore use joint travel together with planned arm motion time to characterize the deployment cost.
Joint travel is computed by summing the absolute changes of all seven joint angles over consecutive samples of the certified execution trajectories.
It includes Home entry and return motions and any certified Home fallback connections, while rail translation and lift adjustment are excluded from this arm-specific metric.

The reported number of active views counts only active RGB-D acquisitions; observations collected during the passive sweep are excluded.

\subsection{Reachability-Constrained Execution}
In this section, we introduce robot-specific reachability and motion constraints for executing our one-shot view plans on the greenhouse platform.
These constraints restrict the candidate set available for view selection without changing the set-cover formulation.
We describe candidate filtering, trajectory certification, and the execution procedure used for active acquisition.

\subsubsection{Candidate Reachability}
A candidate camera pose inside the sampling workspace does not necessarily correspond to a feasible configuration of the deployed manipulator.
We therefore introduce collision-aware inverse kinematics (IK) filtering before view selection.
Each candidate is assigned to a rail station according to its position along the row and evaluated at the two lift states.
The calibrated end-effector-to-camera transformation, together with the rail-and-lift configuration, converts the candidate camera pose into an end-effector target for IK.
A candidate is retained only if a joint configuration satisfying the joint limits and collision constraints is found.

Collision checking uses the robot model together with the platform, the attached camera assembly, and keep-out volumes representing the neighboring crop rows.
The camera assembly includes the camera body, its mounting structure, and the USB connector, allowing these components to be considered during collision checking.
When a candidate is feasible at both lift states, the upper state is preferred, and one joint configuration is retained for subsequent processing.
This stage establishes endpoint feasibility; the connecting motions are certified separately.

\begin{figure*}[!t]
\centering
\includegraphics[width=\textwidth]{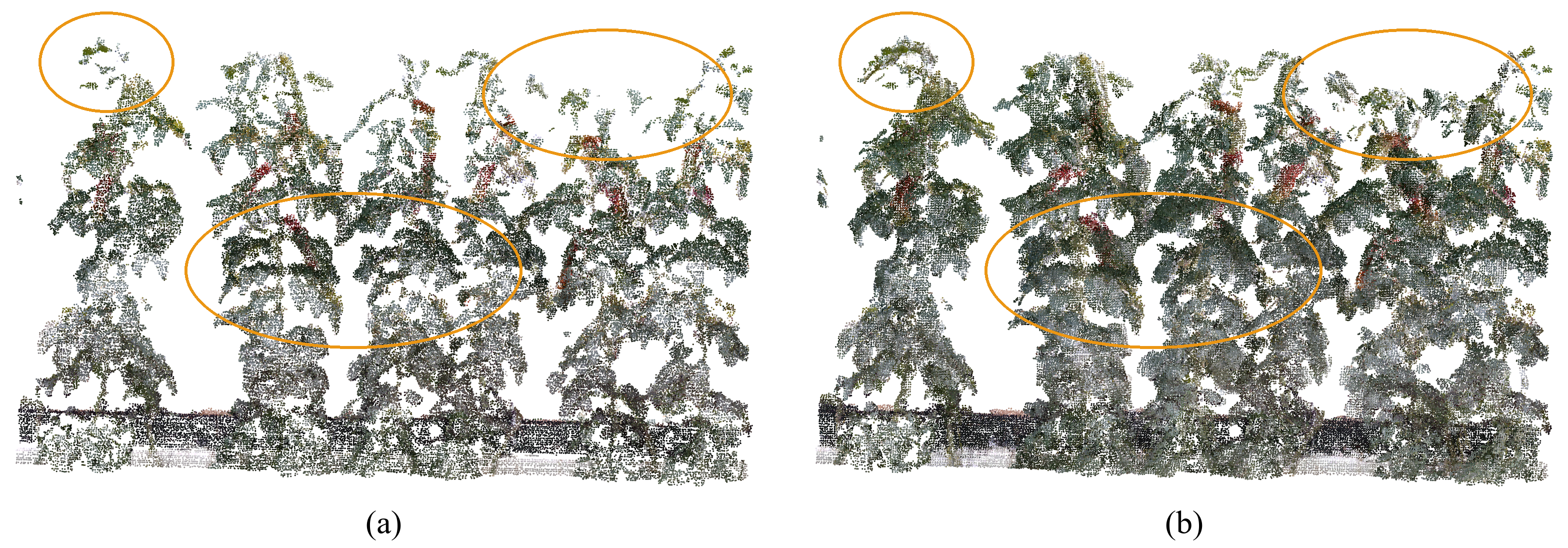}
\caption{
Qualitative reconstruction comparison from an additional real-world acquisition using the Case 2 view plan, recorded for the Supplementary Video.
(a) Passive reconstruction.
(b) Reconstruction combining passive and active observations.
Both reconstructions are shown from the same viewpoint and at the same scale.
Paired highlights mark corresponding regions in the two reconstructions.
Active acquisition supplements the passive sweep with additional geometry in the upper and central canopy.
}
\label{fig:real_world_reconstruction}
\end{figure*}

\subsubsection{Home--View--Home Trajectory Certification}
Endpoint IK feasibility alone is insufficient for execution because the external camera cable also constrains the intervening arm motion.
A trajectory may violate cable-related joint limits or coupled wrist constraints even when its endpoint is feasible.
We therefore certify a Home--View--Home connection for each remaining candidate before view selection.

We use a fixed Home configuration with all seven joint angles set to zero.
For each candidate, MoveIt~\cite{chitta2016ros} with RRTConnect plans an outbound trajectory from Home to the retained joint configuration, with velocity and acceleration scaling factors both set to 0.1.
The trajectory is checked against hardware joint limits, additional cable-related joint bounds, and the collision scene.
Cable-related constraints comprise Home-centered soft joint limits and an empirically determined coupling rule for joints 6 and 7 of the xArm7: when joint 6 is below $-70^\circ$, joint 7 must not exceed $120^\circ$.
These empirical limits depend on the camera cable routing and mounting arrangement and are determined for our deployed configuration.
The constraints are enforced along the densified trajectory.
Dense collision checking uses a maximum per-joint sampling step of 0.01\,rad and requires a minimum self-clearance of 12\,mm.
The return trajectory is constructed by exact time reversal of the validated outbound trajectory, followed by endpoint and velocity/acceleration checks.
Only candidates passing this certification are made available to view selection.
This certification ensures that every retained candidate has a validated Home--View--Home connection under the modeled constraints, allowing subsequent view selection to operate on an executable candidate set without changing its optimization formulation.

The certified Home connections provide validated approach and retreat paths and support the fallback connections used during route execution.
They do not require the robot to return to Home after every observation; direct connections between consecutive selected views are certified separately.

\subsubsection{View Selection from Certified Candidates}
We apply the original set-cover-based view selection to the certified candidate set, using a covering resolution of 4\,cm.
The robot-specific constraints restrict the available candidates without changing the optimization objective or the coverage formulation.
The coverage demand consists of surface voxels observable from at least one certified candidate, excluding those already covered by the passive observations.
We solve a single global set-cover problem over candidates from all eight acquisition blocks.
Visibility evaluation is accelerated using CUDA, while set-cover optimization is performed by Gurobi on the CPU.

\subsubsection{Route Certification and Execution}
The passive sweep ends near the farthest rail station with the lift in its lower state.
We therefore execute the lower blocks in station order $3 \rightarrow 2 \rightarrow 1 \rightarrow 0$, raise the lift at station 0, and execute the upper blocks in order $0 \rightarrow 1 \rightarrow 2 \rightarrow 3$.
This order limits additional rail repositioning and requires only one transition from the lower to the upper lift state during active acquisition.

Integrating view planning with robot motion planning can improve the physical efficiency of active acquisition~\cite{zaenker2023graph,jose2025go}.
To account for robot-specific motion beyond the Euclidean camera-space distance used in simulation, we order the selected views within each block using a joint-space open TSP starting from Home.
The edge cost is the sum of absolute joint-angle differences between the retained endpoint configurations, without angular wrapping.
Within each block, we solve the TSP globally using the same procedure as in simulation, replacing only the camera-space edge cost with the joint-space cost.
The TSP has a free final viewpoint; the subsequent return to Home is included in the execution trajectories and reported motion costs.

After ordering and before active acquisition, MoveIt~\cite{chitta2016ros} plans and certifies direct trajectories between consecutive selected views, translating the viewpoint sequence into executable robot motions.
If a direct connection cannot be certified, it is replaced by a View--Home--View connection using the previously certified Home trajectories.
The arm starts and ends each block at Home, and rail or lift repositioning is performed with the arm at this configuration.

During acquisition, the robot replays the certified trajectories while the rail and lift remain fixed within each block.
At each viewpoint, the controller first verifies that the arm has reached the target joint configuration.
The robot then waits for 2\,s before acquiring a fresh synchronized RGB-D observation, excluding measurements buffered before settling.
The view ID, timestamps, actual arm pose, rail odometry, and lift state are recorded with each observation for subsequent reconstruction.

\begin{table}[!t]
\centering
\resizebox{\columnwidth}{!}{%
\begin{tabular}{|c|c|c|c|c|}
\hline
Case & \makecell[c]{Raw\\Candidates} & \makecell[c]{IK-Reachable\\Candidates} & \makecell[c]{Home-Certified\\Candidates} & \makecell[c]{Planned\\Active Views} \\
\Xhline{1.2pt}
1 & 800 & 631 & 555 & 171 \\
\hline
2 & 800 & 638 & 564 & 173 \\
\hline
3 & 800 & 638 & 560 & 177 \\
\Xhline{1.2pt}
\end{tabular}
}
\caption{
Candidate filtering and active-view selection under robot-specific execution constraints.
IK-reachable candidates satisfy endpoint feasibility checks, while Home-certified candidates additionally pass Home--View--Home trajectory certification.
The planned active views are selected by the original set-cover formulation at a covering resolution of 4\,cm.
Across the three cases, certification retains 555--564 of the 800 raw candidates, from which one-shot planning selects 171--177 active views.
}
\label{tab:reachability_filtering}
\end{table}

\begin{table*}[!t]
\centering
\renewcommand{\arraystretch}{1.15}
\resizebox{0.9\textwidth}{!}{%
\begin{tabular}{|c|c|c|c|c|c|c|}
\hline
\multirow{2}{*}{\makecell[c]{Case\\(Session Date)}} &
\multirow{2}{*}{\makecell[c]{Previous-Session\\Input (Date)}} &
\multicolumn{2}{c|}{Number of Active Views} &
\multirow{2}{*}{\makecell[c]{Passive\\Occupied Voxels}} &
\multirow{2}{*}{\makecell[c]{Additional Occupied Voxels\\(OCC Gain)}} &
\multirow{2}{*}{\makecell[c]{Planned Arm\\Joint Travel (rad)}} \\
\cline{3-4}
& & Planned & Executed & & & \\
\Xhline{1.2pt}
\makecell[c]{Case 1\\(Sep. 30)} &
\makecell[c]{Dense Bootstrap\\(Sep. 29)} &
171 & 171 & 149,576 & 70,176 (46.92\%) & 1,355.96 \\
\hline
\makecell[c]{Case 2\\(Oct. 2)} &
\makecell[c]{Previous Ours\\(Sep. 30)} &
173 & 173 & 140,879 & 68,348 (48.52\%) & 1,180.40 \\
\hline
\makecell[c]{Case 3\\(Oct. 5)} &
\makecell[c]{Previous Ours\\(Oct. 2)} &
177 & 177 & 138,745 & 70,807 (51.03\%) & 1,193.14 \\
\Xhline{1.2pt}
\end{tabular}
}
\caption{
Real-world acquisition and reconstruction results across three monitoring cases.
All dates refer to 2026.
Dates in the previous-session input column indicate when the input reconstructions were acquired.
Planned and executed view counts include only active acquisitions.
Additional occupied voxels are reported with the relative OCC gain in parentheses.
Occupancy is evaluated using a common 10\,mm voxel grid, the same Plant OBB, and identical outlier filtering.
All planned active views are successfully executed, and active acquisition increases occupied-voxel counts by 46.92--51.03\% over the respective passive baselines.
}
\label{tab:real_world_results}
\end{table*}

\begin{table}[!t]
\centering
\resizebox{0.75\columnwidth}{!}{%
\begin{tabular}{|c|c|c|c|}
\hline
Block &
\makecell[c]{Lift State /\\Station} &
\makecell[c]{Active\\Views} &
\makecell[c]{Occupied\\Voxels} \\
\Xhline{1.2pt}
1 & Low / 3  & $28.3 \pm 2.3$ & $23,642 \pm 1,520$ \\
\hline
2 & Low / 2  & $24.7 \pm 3.1$ & $23,810 \pm 4,557$ \\
\hline
3 & Low / 1  & $28.0 \pm 4.4$ & $26,726 \pm 4,984$ \\
\hline
4 & Low / 0  & $20.7 \pm 2.5$ & $14,530 \pm 2,160$ \\
\Xhline{1.2pt}
5 & High / 0 & $17.0 \pm 1.0$ & $11,099 \pm 582$ \\
\hline
6 & High / 1 & $17.3 \pm 4.0$ & $12,959 \pm 1,821$ \\
\hline
7 & High / 2 & $20.3 \pm 4.5$ & $15,097 \pm 4,588$ \\
\hline
8 & High / 3 & $17.3 \pm 1.5$ & $9,297 \pm 1,358$ \\
\Xhline{1.2pt}
\multicolumn{2}{|c|}{Total / Union} &
$173.7 \pm 3.1$ & $124,457 \pm 3,107$ \\
\Xhline{1.2pt}
\end{tabular}
}
\caption{
Block-wise active-view allocation and acquired occupancy.
Values are mean $\pm$ sample standard deviation across the three monitoring cases.
Occupancy is counted on a common 10\,mm grid using workspace-cropped, ICP-registered active clouds, without additional final SOR filtering.
In the last row, active-view counts are summed across blocks, whereas occupied voxels are counted from their union rather than by summing individual block counts.
These totals and unions are computed separately for each monitoring case before calculating the reported mean and standard deviation.
Occupancy includes overlap with the passive map and does not represent net reconstruction gain.
View allocation is nonuniform, with more active views assigned to the lower lift blocks on average.
}
\label{tab:block_allocation}
\end{table}

\begin{table}[!t]
\centering
\renewcommand{\arraystretch}{1.15}
\resizebox{\columnwidth}{!}{%
\begin{tabular}{|c|l|c|}
\hline
Stage & Component & Time (s) \\
\Xhline{1.2pt}
\multirow{2}{*}{\makecell[c]{Passive\\mapping}}
& Passive acquisition & $33.356 \pm 0.440$ \\
& Passive reconstruction & $18.899 \pm 2.780$ \\
\hline
\multirow{5}{*}{\makecell[c]{Candidate\\preparation}}
& Nonrigid registration & $4.228 \pm 1.863$ \\
& Transferred-target generation & $0.197 \pm 0.046$ \\
& Candidate sampling & $3.859 \pm 0.446$ \\
& Endpoint feasibility checks & $26.500 \pm 1.099$ \\
& Home--View--Home certification & $543.408 \pm 14.357$ \\
\hline
\multirow{2}{*}{\makecell[c]{View\\selection}}
& CUDA visibility evaluation & $0.626 \pm 0.175$ \\
& Set-cover optimization & $0.597 \pm 0.193$ \\
\hline
\multirow{2}{*}{\makecell[c]{Route\\planning}}
& Joint-space TSP & $0.048 \pm 0.029$ \\
& Inter-view trajectory certification & $83.989 \pm 3.833$ \\
\hline
\multirow{2}{*}{\makecell[c]{Active\\acquisition}}
& Actual acquisition span & $3,507.000 \pm 173.744$ \\
& (Planned arm motion time) & $(1,919.156 \pm 161.276)$ \\
\hline
\multirow{2}{*}{\makecell[c]{Final\\reconstruction}}
& Point-cloud fusion and filtering & $4.357 \pm 0.192$ \\
& Local ICP registration & $0.967 \pm 0.284$ \\
\Xhline{1.2pt}
\multicolumn{2}{|c|}{Reported stage total (s)}
& $4,228.030 \pm 196.938$ \\
\Xhline{1.2pt}
\end{tabular}
}
\caption{
Detailed runtime breakdown of the deployed pipeline.
Values are mean $\pm$ sample standard deviation across the three monitoring cases, rather than repeated timing trials of a fixed input.
Computational timings exclude data loading, writing, and diagnostic exports.
Passive acquisition measures the physical rail sweep.
The actual acquisition span extends from the first completed capture to the last completed capture.
Planned arm motion time is shown in parentheses for reference and is excluded from the reported total; it includes all Home and fallback connections, including initial and final Home motions outside the acquisition span.
It excludes rail/lift motion, settling, and capture.
The reported total sums the defined stage timings rather than measuring complete end-to-end wall-clock turnaround time.
Active acquisition dominates the reported runtime, while trajectory certification accounts for most planning computation.
}
\label{tab:real_world_runtime}
\end{table}

\subsection{Deployment Results and Analysis}

In this section, we evaluate the execution feasibility, reconstruction gains, and computational and motion costs of the deployed system.
We first report acquisition completion and occupancy gains across the three monitoring cases, followed by block-wise view allocation and acquired observations.
We then analyze stage-wise runtime and examine how CUDA-accelerated visibility evaluation and joint-space route ordering improve computational and motion efficiency, respectively.

\subsubsection{Acquisition and Reconstruction Results}
Table~\ref{tab:reachability_filtering} reports candidate filtering and view-selection results, while Table~\ref{tab:real_world_results} summarizes acquisition completion, reconstruction gains, and arm motion costs across the three monitoring cases.

\textbf{Execution feasibility.}
Of the 800 raw candidates in each case, 631--638 satisfy endpoint feasibility checks, and 555--564 pass Home--View--Home trajectory certification.
The planner selects 171, 173, and 177 active views in Cases 1--3, respectively.
All planned views are successfully executed across all eight acquisition blocks without view replanning during acquisition.
The agreement between planned and executed view counts demonstrates the execution feasibility of the certified plans across the tested cases.
Planned arm joint travel ranges from 1,180.40 to 1,355.96\,rad; the influence of route ordering is analyzed subsequently.

\textbf{Reconstruction gains from active acquisition.}
Active acquisition adds 68,348--70,807 occupied voxels across the three cases, corresponding to OCC gains of 46.92\%, 48.52\%, and 51.03\% over their respective passive baselines.
The absolute increases are similar, while the relative gains also depend on the passive baseline size.
Fig.~\ref{fig:real_world_reconstruction} provides a qualitative illustration from an additional real-world acquisition using the Case 2 view plan.
Compared with the passive reconstruction in Fig.~\ref{fig:real_world_reconstruction}(a), the reconstruction incorporating active observations in Fig.~\ref{fig:real_world_reconstruction}(b) contains additional geometry in the upper and central canopy, as illustrated by the paired highlighted regions.
The acquisition process and reconstruction comparison are also presented in the Supplementary Video\footnote{\url{https://youtu.be/THCvIye3yx4}}.

\textbf{Practical applicability to repeated monitoring.}
The monitoring sequence is initialized once using a dense bootstrap reconstruction, after which Cases 2 and 3 reuse reconstructions produced by our method in preceding sessions.
Both subsequent cases obtain additional reconstructed geometry without repeating dense acquisition.
This observation is qualitatively consistent with the source-substitution results in Table~\ref{tab:previous_session_source}, which support using a previous planned reconstruction as the input for subsequent monitoring.
Case 3 also obtains additional geometry following routine harvesting.
Such changes are also represented in the real-world greenhouse-row dataset underlying the simulation-based evaluation, which includes inter-session changes associated with plant growth and routine harvesting.
Together, these results support repeated monitoring with a one-time dense initialization followed by previous-session-guided active acquisition under the tested conditions.

\subsubsection{Coverage-Driven View Allocation}
Table~\ref{tab:block_allocation} shows how the global set-cover solution distributes active views across the eight rail-and-lift blocks.
The allocation follows the coverage requirements of the planning geometry, without imposing a fixed view budget on individual blocks.
Denser or more occluded regions can require more complementary views to cover their observable geometry.
This reflects the role of the set-cover formulation: views are selected jointly across blocks according to their contributions to covering the planning demand, rather than being distributed uniformly.

The corresponding acquired occupancy is consistent with this nonuniform allocation.
For example, lower blocks 1 and 3 receive 28.3 and 28.0 active views on average and observe 23,642 and 26,726 occupied voxels, respectively.
In comparison, upper blocks 5 and 8 receive 17.0 and 17.3 views and observe 11,099 and 9,297 occupied voxels.
These examples show that larger view allocations accompany larger amounts of acquired plant geometry, while the jointly selected views provide observations along the row and across plant heights.

\subsubsection{Stage-Wise Runtime Analysis}
Table~\ref{tab:real_world_runtime} reports the runtime of the deployed pipeline, with individual components grouped according to the monitoring workflow.
We report mean and sample standard deviation across the three monitoring cases.

\textbf{Planning computation is dominated by trajectory certification.}
Home--View--Home certification takes $543.41$\,s on average, and certification of the ordered inter-view connections requires a further $83.99$\,s.
In comparison, nonrigid registration takes $4.23$\,s, CUDA visibility evaluation and set-cover optimization together take $1.22$\,s, and joint-space TSP takes $0.048$\,s on average.
Thus, most planning computation is spent establishing executable robot motions under the modeled constraints, rather than selecting or ordering viewpoints.
We use MoveIt as an existing motion-planning backend to support deployment; optimizing robot trajectory generation is beyond the scope of this work, which focuses on previous-session reconstruction-guided view planning.
GPU-accelerated motion generation, such as cuRobo~\cite{sundaralingam2023curobo}, could be investigated to reduce this deployment cost while retaining the required collision and cable-constraint checks.

\begin{table}[!t]
\centering
\renewcommand{\arraystretch}{1.15}
\resizebox{0.9\columnwidth}{!}{%
\begin{tabular}{|l|c|c|}
\hline
Component & CPU Pipeline (s) & CUDA Pipeline (s) \\
\Xhline{1.2pt}
Visibility evaluation
& $223.430 \pm 11.244$
& $0.626 \pm 0.175$ \\
\hline
Set-cover optimization
& $0.460 \pm 0.033$
& $0.597 \pm 0.193$ \\
\Xhline{1.2pt}
Total view selection
& $223.890 \pm 11.277$
& $1.224 \pm 0.299$ \\
\Xhline{1.2pt}
\end{tabular}
}
\caption{
Effect of CUDA-accelerated visibility evaluation on view-selection runtime.
Values are mean $\pm$ sample standard deviation across three cases, with one run per pipeline per case.
Both pipelines use the same CPU set-cover implementation, with timings including initialization, model construction, optimization, and associated processing.
CUDA accelerates visibility evaluation, reducing mean total view-selection time from 223.890\,s to 1.224\,s.
}
\label{tab:cuda_visibility}
\end{table}

\begin{table}[!t]
\centering
\renewcommand{\arraystretch}{1.15}
\resizebox{\columnwidth}{!}{%
\begin{tabular}{|c|c|c|c|}
\hline
Metric &
\makecell[c]{Home--View\\--Home} &
\makecell[c]{3D-space\\TSP} &
\makecell[c]{Joint-space\\TSP} \\
\Xhline{1.2pt}
\makecell[c]{Planned arm\\motion time (min)} &
$73.78 \pm 1.87$ &
$44.47 \pm 1.43$ &
$31.99 \pm 2.69$ \\
\hline
\makecell[c]{Arm joint\\travel (rad)} &
$3,201.05 \pm 44.21$ &
$1,825.02 \pm 62.57$ &
$1,243.17 \pm 97.89$ \\
\hline
\makecell[c]{Camera XYZ\\visitation length (m)} &
--- &
$26.66 \pm 1.10$ &
$52.13 \pm 2.46$ \\
\hline
\makecell[c]{Home fallback\\connections} &
--- &
$15.7 \pm 3.5$ &
$9.7 \pm 5.5$ \\
\Xhline{1.2pt}
\end{tabular}
}
\caption{
Effect of route ordering on certified arm motion costs.
Values are mean $\pm$ sample standard deviation across the three monitoring cases.
Arm motion costs include Home and fallback connections and exclude rail/lift motion, settling, and capture.
Camera XYZ visitation length sums straight-line distances between consecutive observation positions within blocks, excluding Home and inter-block transitions.
Fallback connections replace inter-view motions that cannot be directly certified.
Joint-space ordering reduces arm motion costs despite a longer camera-position visitation length.
}
\label{tab:route_ordering}
\end{table}

\textbf{Acquisition time reflects conservative execution settings.}
The actual active acquisition span averages $58.45$\,min, with a planned arm motion time of $31.99$\,min provided as a reference.
The planned arm trajectories use conservative velocity and acceleration scaling factors of 0.1 to support safe experimental deployment.
The actual acquisition span additionally includes a 2\,s settling period before each RGB-D acquisition, rail translation, lift repositioning, and capture and control overhead.
If faster monitoring is required, these engineering settings could be optimized through further validation of motion safety and observation quality.
The reported acquisition time therefore characterizes the current conservative deployment configuration rather than an intrinsic timing requirement of the view-planning method.

\subsubsection{Effect of CUDA-Accelerated Visibility Evaluation}
We compare the multithreaded CPU and CUDA implementations of visibility evaluation using the same case inputs and certified candidate sets.
Both pipelines use the same set-cover formulation and Gurobi solver on the CPU.
Table~\ref{tab:cuda_visibility} reports component-wise timings and total view-selection time across the three monitoring cases.
Absolute runtimes are not directly comparable with those of the simulation-based evaluation because the hardware, scene geometry, and candidate-view counts differ.

Visibility evaluation dominates view-selection time in the CPU pipeline.
CUDA accelerates this component by approximately $300$--$500\times$, yielding an overall view-selection speedup of approximately $150$--$260\times$ across the three cases.
The acceleration comes from visibility evaluation, while set-cover optimization retains the same formulation and CPU solver.
Both pipelines return optimal solutions with matching selected-view counts in all three cases and zero uncovered demand under their respective computed visibility sets.

\subsubsection{Effect of Route Ordering}
We compare Home--View--Home execution, 3D-space TSP, and joint-space TSP using the same selected views, endpoint joint configurations, and block order.
Each strategy is evaluated through trajectory planning and certification under the same execution constraints.

As shown in Table~\ref{tab:route_ordering}, joint-space TSP reduces planned arm motion time by approximately 28\% relative to 3D-space TSP and 57\% relative to Home--View--Home, averaged over the three paired cases.
The corresponding reductions in joint travel are approximately 32\% and 61\%, respectively.
Although 3D-space TSP produces a shorter camera-position visitation length, nearby camera positions can require substantial joint motion, and their connecting trajectories must satisfy collision and cable constraints.
Joint-space ordering better reflects these manipulator-specific motion costs, supporting the replacement of the camera-space edge cost while retaining the original select-then-order pipeline.

\subsubsection{Summary}
Overall, the real-world experiments lead to three main takeaways.
First, robot-specific candidate filtering and trajectory certification enable successful execution of all planned views across the tested cases while retaining the original set-cover formulation.
Second, a one-time dense bootstrap followed by reuse of robot-acquired reconstructions supports repeated monitoring, with active acquisition consistently adding geometry beyond the passive warm start, including after routine harvesting.
Third, CUDA-accelerated visibility evaluation reduces view-selection latency, while joint-space route ordering reduces arm motion costs compared with camera-space ordering.
The main measured deployment costs lie in trajectory certification and physical acquisition under the conservative execution settings.
Together, these results support the practical deployment of previous-session reconstruction-guided one-shot view planning under the tested greenhouse conditions.

\section{Discussion}

\subsubsection{Deployment Considerations}
The real-world implementation illustrates how platform-specific constraints can be incorporated through candidate filtering and trajectory certification while retaining the original set-cover formulation.
The joint limits, cable restrictions, and rail-and-lift configurations used here depend on the deployed platform and sensing arrangement.
These constraints determine the available candidate set and therefore influence the geometry that can be covered by active acquisition.
Applying the method to another platform would require adapting its feasibility checks and motion-cost model.
The reported acquisition time also reflects conservative motion and settling settings, which could be optimized through further validation of execution safety and observation quality.

\subsubsection{Broader Applicability}
Although we focus on greenhouse crop rows, the proposed formulation is not limited to this environment.
The individual-plant setting reflects a generic object-centric repeated reconstruction problem and can naturally extend to field robotics where plants are revisited over time, such as ground robots scanning individual crops in open fields or aerial platforms monitoring tree canopies.
In such scenarios, reusing a previous-session reconstruction can similarly reduce redundant acquisition by focusing observations on under-observed structures while accounting for execution effort.

\subsubsection{Future Directions}
Our current system uses warm-start reconstructions, obtained through passive mapping for rows and a short NBV warm-up for individual plants, to support cross-session reconstruction transfer.
Incorporating semantic and temporal consistency could further strengthen correspondence and view planning under occlusion, growth, and harvesting.
Semantic landmarks, such as fruits, peduncles, and stem junctions, could provide additional cues for cross-session association and help distinguish persistent structures from newly appearing or removed organs.
These associations could also guide view planning toward persistently under-observed organs relevant to monitoring tasks, extending the current geometry-based coverage objective toward organ-specific information acquisition.

\subsubsection{Agricultural Relevance}
Repeated 3D reconstruction provides a geometric basis for tracking plant development across monitoring sessions.
When combined with downstream semantic analysis and temporal association, these reconstructions could support growth assessment, yield estimation from fruit counts and sizes, and detection of abnormal changes in plant structure.
Our approach supports these applications by reducing repeated acquisition effort while maintaining reconstruction coverage across monitoring sessions.

\section{Conclusion}
Efficient view planning is a key enabler for repeated plant monitoring, where 3D reconstructions need to be acquired over multiple sessions under changing plant geometry.
This paper presents a one-shot view-planning approach guided by a previous-session reconstruction, which transfers the previous-session geometry to the current session via warm-started non-rigid registration and conservative inflation, and then performs set-cover-based view selection with global path planning.
Experiments on real plants captured over repeated monitoring sessions, including public single-plant scans and a newly collected greenhouse crop-row dataset, evaluate both efficiency and robustness.
Real-world greenhouse experiments further demonstrate practical deployment under robot-specific execution constraints.
The results support the use of previous-session reconstructions as an actionable geometric approximation for repeated plant monitoring: robust geometry transfer coupled with one-shot set selection and global routing provides an efficient alternative to purely session-local planning, and highlights how the most effective planning variables depend on the monitoring setting.
Efficient repeated reconstruction can support tracking crop development over time and provide a basis for informed agricultural management.

\bibliographystyle{IEEEtran}
\footnotesize
\bibliography{tfr2026}

\end{document}